\documentclass{article}

\usepackage[final]{neurips_2025_ai4science}

\usepackage[utf8]{inputenc} 
\usepackage[T1]{fontenc}    
\usepackage{hyperref}       
\usepackage{url}            
\usepackage{booktabs}       
\usepackage{amsfonts}       
\usepackage{nicefrac}       
\usepackage{microtype}      
\usepackage{xcolor}         
\usepackage{pifont}
\usepackage{makecell}
\usepackage[table]{xcolor}
\usepackage{float}

\usepackage{graphicx} 
\usepackage{caption}
\usepackage{wrapfig}

\usepackage{mathrsfs}
\usepackage{amsmath}
\usepackage{amsthm}
\usepackage{amssymb}
\usepackage{tablefootnote}
\usepackage{multirow}
\usepackage{enumerate}
\usepackage{color}
\usepackage{xcolor}
\usepackage{dsfont}

\definecolor{LightCyan}{rgb}{0.88,1,1}

\usepackage{algorithm}
\usepackage{algpseudocode}
\allowdisplaybreaks[4]
\usepackage{bm,todonotes}
\allowdisplaybreaks

\newcounter{subassumption}[asu]

\makeatletter
\renewcommand{\p@subassumption}{\theasu}
\makeatother

\newtheoremstyle{remarkstyle}
  {}                    
  {}                    
  {\normalfont}         
  {}                    
  {\itshape}            
  {.}                   
  { }                   
  {}                    
\theoremstyle{remarkstyle}

\usepackage{amsmath,amsfonts,bm}

\def\1{\bm{1}}

\DeclareMathAlphabet{\mathsfit}{\encodingdefault}{\sfdefault}{m}{sl}
\SetMathAlphabet{\mathsfit}{bold}{\encodingdefault}{\sfdefault}{bx}{n}

\renewcommand{\xi}{\zeta}

\def\0{{\bf 0}}
\def\1{{\bf 1}}

\title{
Improved Therapeutic Antibody Reformatting through Multimodal Machine Learning
}

\author{Jiayi Xin\thanks{Work done while at BigHat Biosciences}\\
University of Pennsylvania\\
\texttt{jiayixin@seas.upenn.edu}
\And
Aniruddh Raghu\\
BigHat Biosciences\\
\texttt{araghu@bighatbio.com}
\And
Nick Bhattacharya\\
BigHat Biosciences\\ 
\And 
Adam Carr\\
BigHat Biosciences\\ 
\And
Melanie Montgomery\\
BigHat Biosciences\\ 
\And
Hunter Elliott\\ 
BigHat Biosciences\\
\texttt{helliott@bighatbio.com}
}

\begin{document}

\maketitle

\begin{abstract}
Modern therapeutic antibody design often involves composing multi-part assemblages of individual functional domains, each of which may be derived from a different source or engineered independently. 
While these complex formats can expand disease applicability and improve safety, they present a significant engineering challenge: the function and stability of individual domains are not guaranteed in the novel format, and the entire molecule may no longer be synthesizable. 
To address these challenges, we develop a machine learning framework to predict \textit{reformatting success} -- whether converting an antibody from one format to another will succeed or not. 
Our framework incorporates both antibody sequence and structural context, incorporating an evaluation protocol that reflects realistic deployment scenarios. 
In experiments on a real-world antibody reformatting dataset, we find the surprising result that large pretrained protein language models (PLMs) fail to outperform simple, domain-tailored, multimodal representations.
This is particularly evident in the most difficult evaluation setting, where we test model generalization to a new starting antibody. In this challenging ``new antibody, no data'' scenario, our best multimodal model achieves high predictive accuracy, enabling prioritization of promising candidates and reducing wasted experimental effort. 

\end{abstract}

\vspace{-10pt}
\section{Introduction}
\vspace{-5pt}
\label{sec:intro}

\begin{wrapfigure}{r}{0.38\textwidth}
\centering
\vspace{-5mm}
\includegraphics[width=0.9\linewidth]{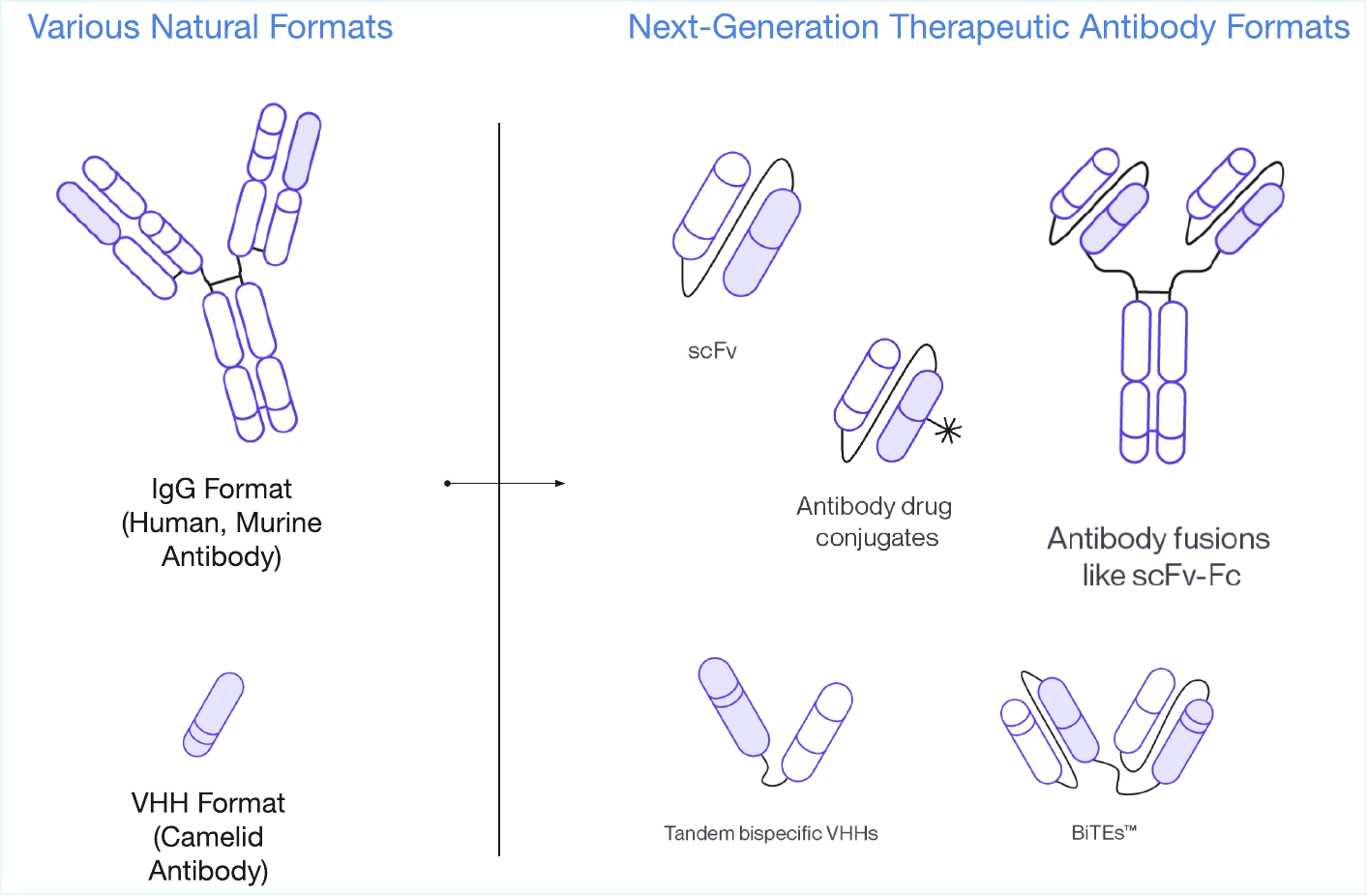}
\vspace{-1mm}
\caption{Examples of natural and engineered antibody formats.}
\label{fig:intro-antibody-formats}
\vspace{-5mm}
\end{wrapfigure}

Antibodies are multi-domain proteins whose architecture underlies their diverse functional roles in the immune system as well as their success as a modern drug modality. In natural antibodies, the variable domains (VH and VL) at the tips mediate target-specific binding, while the constant domains provide structural stability and mediate effector functions \citep{Schroeder2010,Vidarsson2014}. Due to their modularity, modern therapeutic antibodies have been designed in a variety of ``formats'' (Figure \ref{fig:intro-antibody-formats}) that combine these domains in different configurations to achieve specific functional or diagnostic goals \citep{Holliger2005,Dickopf2020}. 

``Reformatting'' refers to converting an antibody from one format to another to enable a novel drug modality or for high-throughput selection, screening, or expression workflows \citep{Schaefer2010}. In this work, we focus specifically on reformatting a natural multichain IgG antibody into single chain variable fragment (scFv), which can allow the binding properties of IgGs to be assayed in high throughput via cell-free expression or phage display, or ease its integration with other functions such as T-cell activation\citep{Hunt2023CFPS,OjimaKato2018Ecobody,Stech2015ReviewCFPS}.

Yet, while antibodies are conceptually modular, reformatting is far from trivial. Properties frequently shift across formats, and binding may be lost upon conversion. The final molecule may also be prone to instability or aggregation, or may not synthesize at all. As a result, reformatting is often a trial-and-error process, with many potential final formats designed and assayed in the wet lab to find workable designs~\citep{Steinwand2014,Schneider2021,Boucher2023,Tu2016,Liu2018,Bailly2020,Jain2017}. 

One promising strategy to reduce the error rate in reformatting is to use a machine learning (ML) model to predict whether a given IgG/scFv pair is likely to reformat successfully, and then reserving \textit{in vitro} validation for designs that are predicted to perform well.  
However, to the best of our knowledge, no prior work has systematically modeled antibody 
reformatting at scale.
Previous work on antibody property prediction has ranged from \emph{developability prediction} (focusing on antibody stability, aggregation, or solubility, using physicochemical features~\citep{lauer2012di}) to supervised models on IgG panels~\citep{hebditch2019peerj, Jain2017}. 
While some models are tailored for specific formats (e.g., scFvs or Fabs), they are not designed to address the unique biophysical shifts introduced by \emph{reformatting}, which can alter stability, aggregation, and even binding~\citep{scfv_fc_igg_frontiers2021, mabs2013_format}. 

In addition, from a machine learning perspective, a key difficulty in predicting reformatting success is the pronounced distribution shift between antibody ``parental families,''. Here we define a parental family as the group of reformatted antibodies derived from a single starting antibody in the source format. While antibodies within a parental family are highly similar, those in other families derived from different starting antibodies can diverge substantially in their sequence-property relationships.

Here, our contribution is to benchmark different representations for predicting reformatting success on a real-world dataset derived from multiple therapeutic development campaigns. We develop a multimodal machine learning framework that predicts IgG$\rightarrow$scFv reformatting success by combining three types of information: \underline{sequence} features, \underline{structure} features, and \underline{biophysical} features. 
We evaluate models under three real-world deployment-motivated scenarios: (1) the \textbf{scFv split}, in which the model predicts outcomes for unseen individual sequences within known families; (2) the \textbf{target-family split}, which simulates fine-tuning on a small set of data from a new family before inference; and (3) the \textbf{parental-family split}, a true zero-shot setting with no training data from a new antibody family.

Our experiments reveal two central findings. First, predicting \emph{protein synthesis failure}---the critical gate that determines whether any downstream assays can even be run---is tractable with multimodal features that integrate sequence, structure, and biophysical representations. In the most challenging \textbf{parental-family split}, which evaluates generalization to entirely unseen antibody families, our linear multimodal model achieves AUROC values above \textbf{88\%}. Second, we find that large pretrained protein language models (PLMs) such as AbLang~\citep{olsen2022ablang}, ISM~\citep{ouyangdistilling}, and DPLM2~\citep{wangdplm} often underperform simple one-hot sequence baselines. This underscores that domain-specific, interpretable representations remain competitive in low-data structural biology settings with strong distribution shifts.

\vspace{-0.4em}
\section{Method}
\vspace{-0.2em}
\label{sec:method}

\subsection{Problem Formulation and Notation}
\label{sec:problem-notation}

Let each antibody be represented by a set of input modalities \(\mathcal{X} = \{\mathbf{x}_{\mathrm{seq}}, \mathbf{x}_{\mathrm{struct}}, \mathbf{x}_{\mathrm{bio}}\}\), where \(\mathbf{x}_{\mathrm{seq}}\) encodes the VH and VL amino acid sequences (over the 20-amino acid alphabet), \(\mathbf{x}_{\mathrm{struct}}\) encodes structure-derived descriptors from predicted 3D structures, and \(\mathbf{x}_{\mathrm{bio}}\) contains biophysical properties. The target variable \(\mathbf{y}\) corresponds either to a protein synthesis success or failure, \(\mathbf{y} \in \{0,1\}\), or a continuous synthesis yield value, \(\mathbf{y} \in \mathbb{R}\) (where higher is better). Given a dataset \(\mathcal{D} = \{(\mathcal{X}_i, \mathbf{y}_i)\}_{i=1}^N\), the objective is to learn a mapping \(f_\theta: \mathcal{X} \mapsto \hat{\mathbf{y}}\) that minimizes a task-appropriate loss \(\mathcal{L}(\mathbf{y}, \hat{\mathbf{y}})\) and generalizes to unseen antibody families, 
i.e., correctly predicting synthesis success for novel families not observed during training.

\subsection{Dataset Construction}
\label{sec:dataset}

\noindent
\textbf{Inputs and targets.}  
Our dataset consists of IgG$\rightarrow$scFv reformatting experiments, where full-length immunoglobulin G (IgG) antibodies are converted into compact single-chain variable fragments (scFvs).  These experiments were conducted across multiple antibody optimization campaigns.  
Each scFv is represented by: (i) VH and VL amino acid sequences, (ii) the linker sequence connecting the domains, (iii) the domain ordering (VH--VL or VL--VH), and (iv) the parental family identifier from which the VH and VL are derived.  
We train separate models for two primary tasks: (1) \emph{protein synthesis outcome classification}, where the target variable $y_{\mathrm{QC}} \in \{0,1\}$ indicates whether a reformatted scFv has synthesized adequately or not, and (2) \emph{yield regression}, where $y_{\mathrm{yield}} \in \mathbb{R}$ measures synthesis yield in ng/$\mu$L.  

\noindent
\textbf{scFv signature and aggregation.}  
We define the \emph{scFv signature} of an input scFv as the tuple:
\[
\textsc{Sig} = (\mathrm{VH}, \mathrm{VL}, \mathrm{linker}, \mathrm{orientation}),
\]
where $\mathrm{VH}$ and $\mathrm{VL}$ are the amino acid sequences of the heavy and light chain variable regions, $\mathrm{linker}$ is the connecting peptide, and $\mathrm{orientation}$ specifies the domain order.  
scFvs sharing the same signature are considered equivalent: their target values are averaged.  
After aggregation, the dataset contains $N=1{,}477$ unique scFv signatures drawn from $56$ parental families across $7$ independent antibody optimization campaigns. Summary statistics are provided in Appendix~\ref{appendix:dataset-statistics}.

\noindent
\textbf{Parental family generalization.}  Within a parental family, scFv sequences are often highly similar, differing by only a few mutations. However, across families, divergence is much greater, creating a strong distribution shift between families. This makes generalization to unseen families a central challenge for our models. This motivates our evaluation strategy, discussed next.

\subsection{Evaluation Protocol}
\label{sec:evaluation}

\begin{wrapfigure}{r}{0.64\textwidth}
\centering
\vspace{-10mm}
\includegraphics[width=0.99\linewidth]{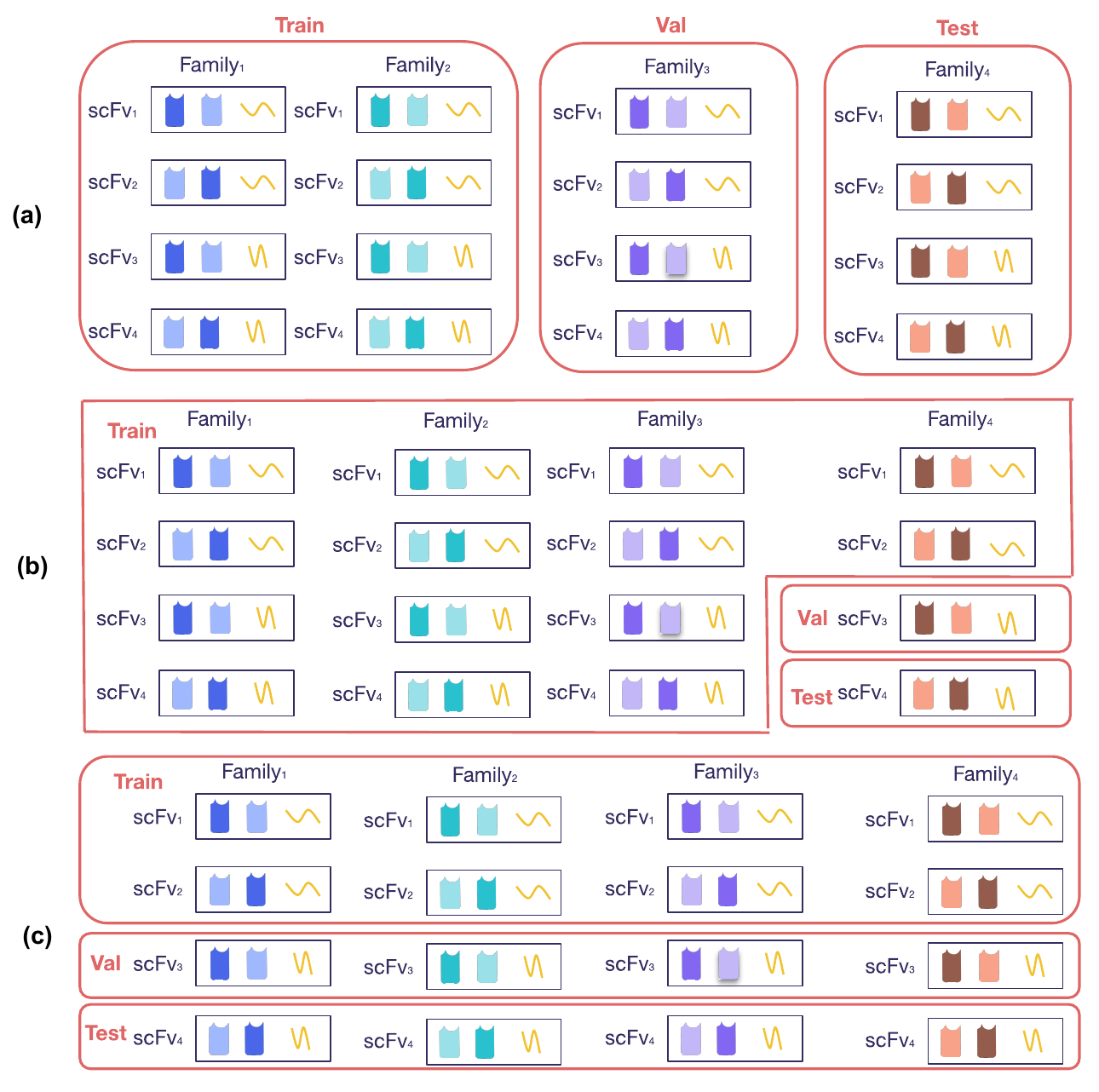}
\vspace{-1mm}
\caption{Three data splits illustrated. (a) Parental Family split. (b) Target Family split. (c) scFv split.}
\label{fig:method-three-splits}
\vspace{-2mm}
\end{wrapfigure}

We evaluate model generalization under three disjoint data partitioning schemes, each corresponding to a realistic deployment scenario (Figure \ref{fig:method-three-splits}): (1) \emph{Parental Family split} (\textsc{New Parental, No Data}): entire parental families are held out from training, forcing zero-shot prediction on novel families with no prior \textit{in vitro} data. (2) \emph{Target Family split} (\textsc{One Batch from New Antibody}): a small batch from the target family is included in training along with data from other families; the remaining scFvs from the target family are used for validation and testing. (3) \emph{scFv split} (\textsc{Lots of Data for All Families}): all families appear in all splits, but individual scFv signatures are unique to a single split, preventing data leakage while maximizing same-family context.

\subsection{Feature Representations}
\label{sec:features}

Our modeling approach integrates three complementary feature modalities: sequence, structure, and biophysical properties. Each modality captures distinct aspects of the scFv--parental IgG relationship, and all features are precomputed and frozen prior to model training.

\paragraph{Sequence-based features.}  
VH and VL amino acid sequences are AHo-aligned\citep{honegger2001aho} to a fixed length ($L_{\mathrm{VL}}=152$, $L_{\mathrm{VH}}=152$) and one-hot encoded. Domain orientation (VH--VL or VL--VH) and linker peptide type are represented as categorical one-hot variables and concatenated with the sequence encodings.  
Given the potential of pretrained sequence models to encode rich information about an input protein, we also evaluate predicting reformatting success using frozen embeddings from two pLMs. First, we consider AbLang~\citep{ablang2_bioinf2024} (heavy/light encoders separately), which is an antibody-specific pLM trained on a large dataset of human antibody sequences. Secondly, we consider ISM~\citep{dreyer2023inverse}, a fine-tuned ESM \citep{rives2021biological} model that is trained to encode structural information in addition to sequence. For both models, residue-level embeddings are computed once and then mean-pooled across the sequence to yield a fixed-length representation. These are kept fixed during downstream training. Our intuition is that such pretrained encoders capture general antibody sequence statistics (e.g., conserved CDR motifs, germline variation) that may aid generalization across families.

\paragraph{Structure-based features.}  
For each unique scFv and its corresponding parental IgG, we predict full-atom 3D structures using Boltz-2~\citep{passaro2025boltz}, a structure prediction model with accuracy comparable to AlphaFold3 across diverse proteins. After performing rigid body alignment between an scFv/IgG pair we compute descriptors from these predictions: (i) the global RMSD of C$\alpha$ atoms between VH/VL domains of the parental IgG and its reformatted scFv, and (ii) per-residue concatenation of aligned parental and scFv C$\alpha$ coordinates with gap indicators, preserving spatial correspondence. 

The motivation is grounded in structural biology, since the function of a protein is intimately related to function, alterations of the structure of individual domains between formats may be related to reformatting success. RMSD serves as a coarse global descriptor (A detailed analysis of RMSD distributions across families is provided in Appendix~\ref{appendix:analysis-structure-per-family}). The per-residue coordinate features offer a more fine-grained view, potentially capturing subtle local rearrangements that influence reformatting outcomes. Representative overlays of VH and VL domains are shown in Figures~\ref{fig:vh_overlay} and~\ref{fig:vl_overlay}.

To benchmark these domain-inspired descriptors against widely used pretrained structural models, we also extract structure-derived embeddings using two models: (i) AbMPNN \citep{dreyer2023inverse} (an antibody inverse folding model that encodes sequences in the context of 3D backbones) and (ii) DPLM2 \citep{wangdplm} (a structure-augmented protein language model). Both provide residue-level embeddings that are mean-pooled to fixed-length vectors for downstream tasks. Our rationale is that inverse folding and structure-augmented pLMs may capture geometric constraints and stability signals that purely sequence-based encoders miss, and thus could help with predicting reformatting success.

\begin{figure*}[t]
  \centering
  \begin{minipage}[t]{0.45\textwidth}
    \centering
    \includegraphics[width=\linewidth]{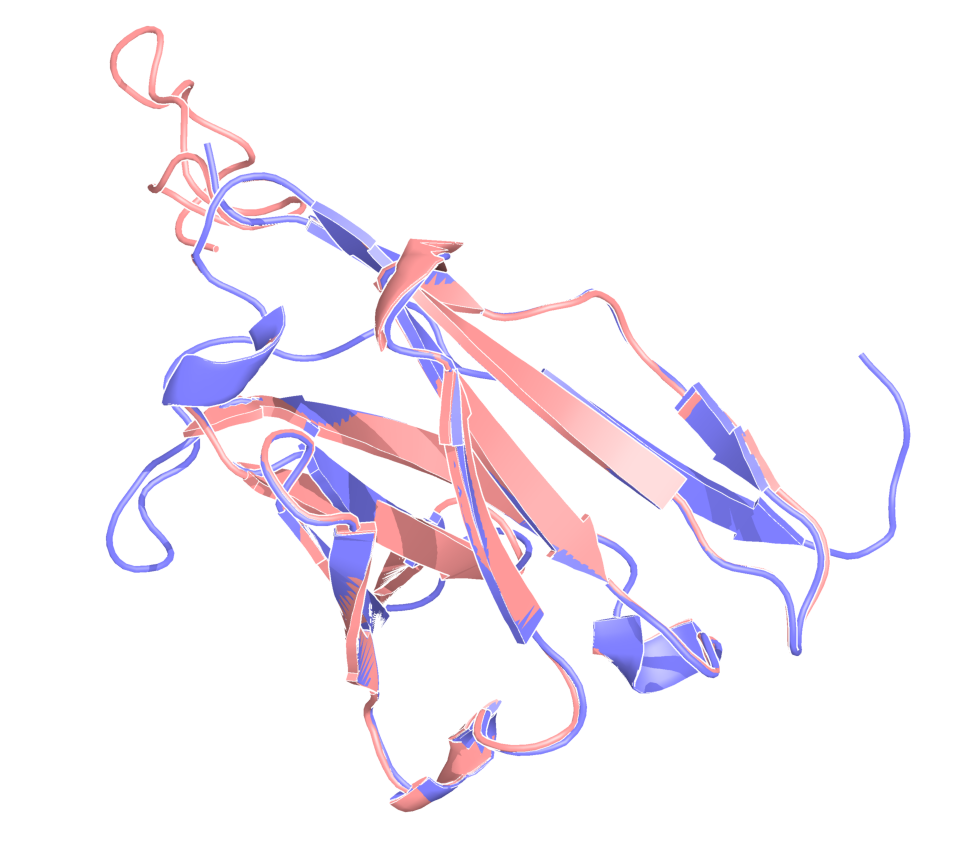}
    \caption{Overlay of predicted VH domain structure between the starting IgG (Purple) and reformatted to scFv (pink). In this example Boltz-2 predicts significant structural alteration upon reformatting.}
    \label{fig:vh_overlay}
  \end{minipage}\hfill
  \begin{minipage}[t]{0.45\textwidth}
    \centering
    \includegraphics[width=\linewidth]{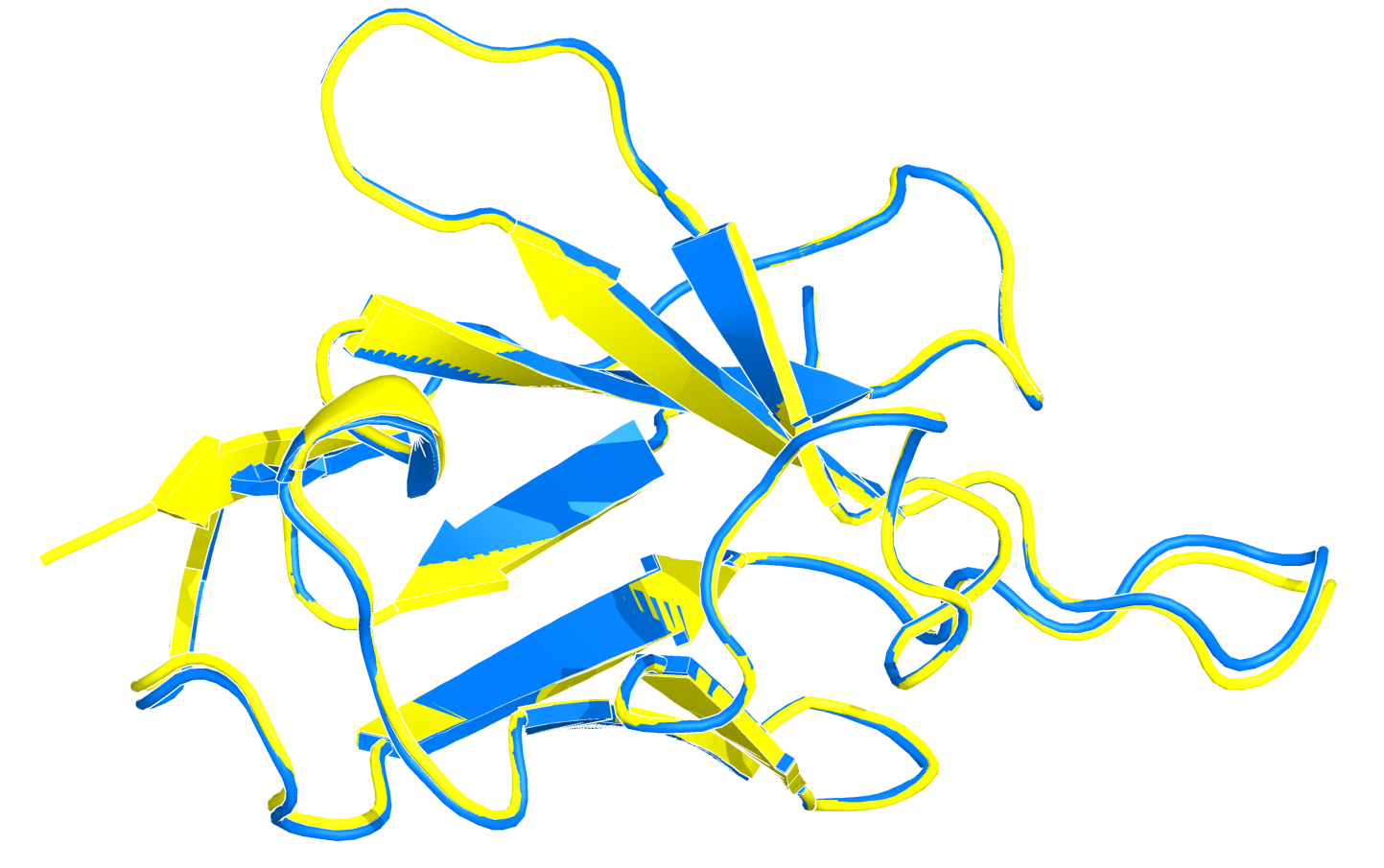}
    \caption{Overlay of predicted VL domain structure between the starting IgG (blue) and reformatted to scFv (yellow).}
    \label{fig:vl_overlay}
  \end{minipage}
\end{figure*}

\paragraph{Biophysical features.}  
From predicted scFv structures, we compute developability metrics using the NaturalAntibody \citep{naturalantibody} platform. Key features derived from the CDR regions include Patch Surface Hydrophobicity (PSH), Patch Negative Charge (PNC), Patch Positive Charge (PPC), and scFv Charge Separation Product (SFvCSP). These are metrics which could be expected to be associated with general antibody stability and expressability. If a score cannot be computed due to modeling failure, the dataset mean is imputed.

\subsection{Model Architectures}
\label{sec:architectures}

\paragraph{Baseline models.}  
We evaluate linear baselines for both protein synthesis outcome classification and yield regression. For classification, we use logistic regression with either L1 or L2 regularization (the choice of regularization is treated as one hyperparameter during tuning); for regression, we use ordinary least squares with optional regularization. Two input configurations are compared: (i) one-hot AHo-aligned VH and VL sequences concatenated with one-hot domain orientation and linker encodings, and (ii) frozen pLM embeddings. All linear models are implemented in \texttt{scikit-learn} and trained with default solvers.

\paragraph{Embedding-based models.}  
We evaluate a fixed-architecture multilayer perceptron (MLP) on frozen pLM embeddings (which outperformed linear models on pLM embeddings). VH and VL embeddings are concatenated prior to the MLP, which contains two hidden layers with ReLU activations and dropout. Hyperparameters such as dropout and learning rate are tuned via grid search, while architecture depth and width remain fixed. Models are trained with AdamW and early stopping.

\paragraph{Multimodal ML models.}  

\begin{wrapfigure}{r}{0.55\textwidth}
\centering
\vspace{-3mm}
\includegraphics[width=0.95\linewidth]{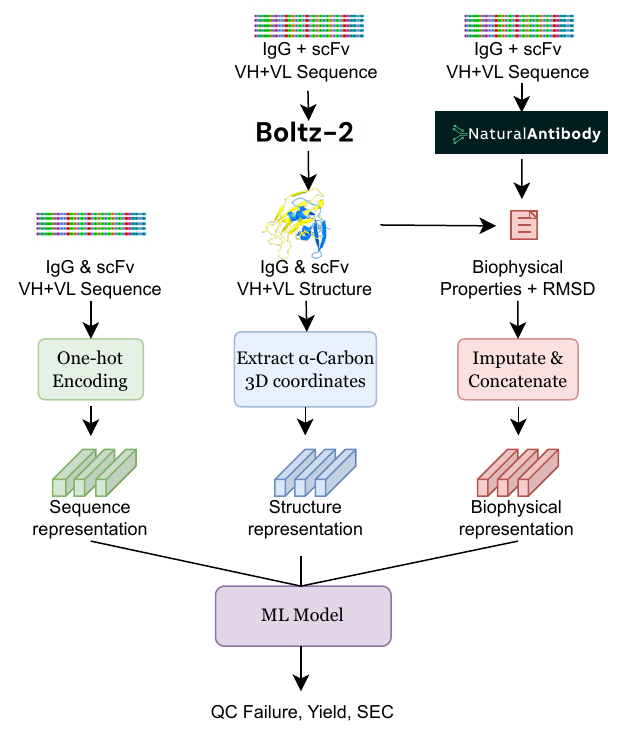}
\vspace{-1mm}
\caption{Multimodal ML pipeline.}
\label{fig:multimodal-ml-pipeline}
\vspace{-3mm}
\end{wrapfigure}

To assess complementarity between modalities, we train simple linear models using combinations of sequence, structure, and biophysical features (Figure \ref{fig:multimodal-ml-pipeline}). 
Hyperparameter tuning follows the same protocol as for the embedding-based models.

\vspace{-0.6em}
\section{Experiments    }
\vspace{-0.1em}
\label{sec:experiment}

\subsection{Experiment Setup}  
Unless otherwise specified, we use an 60\%/10\%/30\% train/validation/test split by the appropriate unit (parental family or scFv signature). For synthesizability classification, we report AUROC and AUPRC in the main text, and accuracy in Appendix. For yield regression, we report Pearson correlation ($r$), and Spearman rank correlation ($\rho$) in Appendix.  Each split type is repeated over $10$ random folds to reduce variance from partitioning; all metrics are reported as mean~$\pm$~standard deviation over folds. All experiments were conducted on a Linux server equipped with 4$\times$NVIDIA A10G GPUs (24\,GB VRAM each) and 12 CPU cores, using CUDA~12.2 and NVIDIA driver version~535.247.01. Hyperparameter search details can be found in Appendix \ref{appendix:detail-hyperparameter}.

\subsection{Simple sequence-only baselines outperform more complex encodings}
\label{sec:seq-perf}
First, we investigate whether simple baselines are sufficient to predict reformatting success, and whether pretrained PLM embeddings offer an advantage over these baselines. In Table~\ref{tab:combined-classification} we compare a logistic regression model, \texttt{LogisticReg}, trained on either one-hot encoded protein sequences or 3D structural coordinates (Section \ref{sec:features}), to sequence-based pLMs (\texttt{AbLang+MLP}, \texttt{ISM+MLP}), a structure-only GNN (\texttt{AbMPNN+MLP}), and a structure-augmented PLM (\texttt{DPLM2+MLP}). We highlight key findings below.

\begin{table*}[!t]
\centering
\caption{\textbf{A linear model with one-hot sequence encoding outperforms pLMs on predicting reformatting synthesis failure.} We compare a logistic regression baseline using either one-hot sequence features (\texttt{vhvl\_only}) or per-residue 3D coordinate features (\texttt{3D\_coord}), to PLM embeddings with an MLP (\texttt{AbLang+MLP}, \texttt{ISM+MLP}), a structure GNN with an MLP (\texttt{AbMPNN+MLP}), a structure-augmented PLM with an MLP (\texttt{DPLM2+MLP}). Best results per column are \underline{\textbf{underlined in bold}}.}
\label{tab:combined-classification}
\resizebox{.9\textwidth}{!}{%
\begin{tabular}{l|l|cc|cc}
\toprule
\textbf{Model} & \textbf{Features} &
\multicolumn{2}{c|}{\textbf{scfv\_signature split}} &
\multicolumn{2}{c}{\textbf{Parental\_Family split}} \\
\cmidrule(lr){3-4} \cmidrule(lr){5-6}
& & \textbf{AUROC} & \textbf{AUPRC} & \textbf{AUROC} & \textbf{AUPRC} \\
\midrule
AbLang+MLP   & \texttt{vhvl\_only}    & 86.35{\scriptsize$\pm$1.39} & 82.15{\scriptsize$\pm$2.26} & 62.58{\scriptsize$\pm$19.37} & 58.75{\scriptsize$\pm$11.70} \\
ISM+MLP      & \texttt{vhvl\_only}    & 80.01{\scriptsize$\pm$1.62} & 74.87{\scriptsize$\pm$2.65} & 58.09{\scriptsize$\pm$9.50}  & 54.02{\scriptsize$\pm$6.10}  \\
DPLM2+MLP    & \texttt{vhvl+struct}   & 79.50{\scriptsize$\pm$1.60} & 73.49{\scriptsize$\pm$2.28} & 47.91{\scriptsize$\pm$5.29}  & 51.21{\scriptsize$\pm$11.13} \\
AbMPNN+MLP   & \texttt{struct\_only}  & 73.38{\scriptsize$\pm$2.11} & 65.89{\scriptsize$\pm$3.89} & 54.30{\scriptsize$\pm$5.69}  & 54.67{\scriptsize$\pm$6.67}  \\
\rowcolor{LightCyan}
LogisticReg  & \texttt{vhvl\_only}    & \underline{\textbf{89.46}}{\scriptsize$\pm$1.63} & \underline{\textbf{87.46}}{\scriptsize$\pm$2.33} & \underline{\textbf{66.35}}{\scriptsize$\pm$10.73} & \underline{\textbf{59.21}}{\scriptsize$\pm$15.65} \\
LogisticReg  & \texttt{3D\_coord}     & 77.00{\scriptsize$\pm$1.00} & 71.00{\scriptsize$\pm$2.00} & 52.00{\scriptsize$\pm$12.00} & 48.00{\scriptsize$\pm$8.00} \\
\bottomrule
\end{tabular}}
\end{table*}

\noindent\ding{182} \textbf{Simple linear models on one-hot encodings outperform embeddings from pLMs and structure encoders.}
We find the surprising result that on the \texttt{scfv\_signature} split, the simple one-hot \texttt{LogisticReg} significantly outperforms more complex embeddings: +\textbf{3.1} AUROC and +\textbf{5.3} AUPRC over AbLang, and nearly +\textbf{10} AUROC / +\textbf{14} AUPRC over DPLM2. Structure-only AbMPNN trails both. The same holds under \texttt{Parental\_Family}, where \texttt{LogisticReg} remains best (e.g.\ +\textbf{3.8} AUROC over AbLang), although performance drops and variance grows substantially. This suggests that the embeddings from these pretrained models do not adequately capture the information required to predict reformatting success.  
Interestingly, we see that a linear model trained on structure features alone does not outperform any model that also incorporates sequence information, but does outperform AbMPNN, the other pure-structure model. This suggests that the 3D coordinate representation might contain useful predictive signal for this task, but is not sufficient as an input representation.

\noindent\ding{183} \textbf{Cross-family generalization remains challenging.}
All unimodal models degrade substantially under \texttt{Parental\_Family}, highlighting the severity of distribution shift across antibody families. We see again that pretrained pLMs and structure-based models underperform relative to the simple linear model baseline. Appendix~\ref{appendix:analysis-structure-per-family} provides additional per-family analysis, showing that the correlation between global RMSD and protein synthesis yield varies widely across parental families, often flipping sign. This heterogeneity helps explain why structure-only features do not generalize well.

\subsection{Multimodal feature encodings improve generalization performance}
\label{sec:multimodal-perf}

\begin{table*}[!t]
\centering
\small
\setlength{\tabcolsep}{4pt} 
\renewcommand{\arraystretch}{1.1} 
\caption{\textbf{Multimodal features consistently outperform sequence-only, with the largest gains under cross-family generalization.} Protein synthesis failure classification using \emph{sequence+structure+biophysics} features (\texttt{multimodal}) vs.\ sequence-only (\texttt{seq\_only}) with a linear classifier. Rows correspond to data splits; columns show head-to-head performance. Values are mean$\pm$std across runs. Best per split/metric is \underline{\textbf{underlined in bold}}.}
\label{tab:main-multimodal-classification}
\begin{tabular}{l
                c c
                c c}
\toprule
\multirow{2}{*}{\textbf{Split}} 
 & \multicolumn{2}{c}{\textbf{AUROC}} 
 & \multicolumn{2}{c}{\textbf{AUPRC}} \\
 & \texttt{multimodal} & \texttt{seq\_only}
 & \texttt{multimodal} & \texttt{seq\_only} \\
\midrule
\texttt{scfv\_signature} 
 & \underline{\textbf{92.93}}{\scriptsize$\pm$3.61} & 89.46{\scriptsize$\pm$1.63}
 & \underline{\textbf{91.18}}{\scriptsize$\pm$4.59} & 87.46{\scriptsize$\pm$2.33} \\
\texttt{Parental\_Family} 
 & \underline{\textbf{88.92}}{\scriptsize$\pm$14.93} & 66.35{\scriptsize$\pm$10.73}
 & \underline{\textbf{85.68}}{\scriptsize$\pm$20.94} & 59.21{\scriptsize$\pm$15.65} \\
\texttt{Fam1} 
 & \underline{\textbf{94.64}}{\scriptsize$\pm$2.39} & 87.68{\scriptsize$\pm$3.91}
 & \underline{\textbf{96.92}}{\scriptsize$\pm$1.53} & 93.23{\scriptsize$\pm$2.15} \\
\texttt{Fam2} 
 & \underline{\textbf{82.96}}{\scriptsize$\pm$9.51} & 71.81{\scriptsize$\pm$7.97}
 & \underline{\textbf{66.10}}{\scriptsize$\pm$13.57} & 54.87{\scriptsize$\pm$9.28} \\
\texttt{Fam3} 
 & \underline{\textbf{93.33}}{\scriptsize$\pm$9.33} & 82.71{\scriptsize$\pm$8.87}
 & \underline{\textbf{97.40}}{\scriptsize$\pm$3.82} & 92.67{\scriptsize$\pm$4.66} \\
\bottomrule
\end{tabular}
\end{table*}

\begin{table*}[!t]
\centering
\small
\setlength{\tabcolsep}{4pt} 
\renewcommand{\arraystretch}{1.1} 
\caption{\textbf{Multimodal features also improve regression, yielding higher Pearson/Spearman correlations in most splits, particularly under cross-family generalization.} Protein yield regression (ng/$\mu$L) using \emph{sequence+structure+biophysics} features (\texttt{multimodal}) vs.\ sequence-only (\texttt{seq\_only}) with a linear regressor. Rows correspond to data splits; columns compare head-to-head results. Values are mean$\pm$std across runs. Best per split/metric is \underline{\textbf{underlined in bold}}.}
\label{tab:main-multimodal-regression}
\begin{tabular}{l
                c c
                c c}
\toprule
\multirow{2}{*}{\textbf{Split}} 
 & \multicolumn{2}{c}{\textbf{Pearson}} 
 & \multicolumn{2}{c}{\textbf{Spearman}} \\
 & \texttt{multimodal} & \texttt{seq\_only} 
 & \texttt{multimodal} & \texttt{seq\_only} \\
\midrule
\texttt{scfv\_signature} 
 & \underline{\textbf{0.641}}{\scriptsize$\pm$0.031} & 0.531{\scriptsize$\pm$0.044}
 & \underline{\textbf{0.741}}{\scriptsize$\pm$0.026} & 0.714{\scriptsize$\pm$0.032} \\
\texttt{Parental\_Family} 
 & \underline{\textbf{0.625}}{\scriptsize$\pm$0.266} & 0.191{\scriptsize$\pm$0.255}
 & \underline{\textbf{0.508}}{\scriptsize$\pm$0.264} & 0.035{\scriptsize$\pm$0.283} \\
\texttt{Fam1} 
 & \underline{\textbf{0.567}}{\scriptsize$\pm$0.035} & 0.550{\scriptsize$\pm$0.061}
 & 0.630{\scriptsize$\pm$0.039} & \underline{\textbf{0.646}}{\scriptsize$\pm$0.067} \\
\texttt{Fam2} 
 & \underline{\textbf{0.288}}{\scriptsize$\pm$0.128} & 0.141{\scriptsize$\pm$0.084}
 & \underline{\textbf{0.207}}{\scriptsize$\pm$0.109} & 0.159{\scriptsize$\pm$0.131} \\
\texttt{Fam3} 
 & \underline{\textbf{0.244}}{\scriptsize$\pm$0.290} & 0.154{\scriptsize$\pm$0.218}
 & 0.241{\scriptsize$\pm$0.220} & \underline{\textbf{0.243}}{\scriptsize$\pm$0.202} \\
\bottomrule
\end{tabular}
\end{table*}

Given the strong performance of the linear model on one-hot encoded features, and the fact that a linear model on the simple 3D coordinate representation outperformed the more complex AbMPNN encoding, we now evaluate whether combining these two modalities together with biophysical descriptors yields improved models that generalize across families.

Tables~\ref{tab:main-multimodal-classification} and Table~\ref{tab:main-multimodal-regression} present results for linear models trained on \texttt{multimodal} features (sequence+structure+biophysics), for the classification and regression tasks respectively. Recall that the \emph{Target Family split} (Section~\ref{sec:evaluation}) mimics a practical setting where only a small batch of measurements is available for a new antibody, and the goal is to extrapolate to the remaining scFvs. This scenario is particularly relevant for antibody engineering pipelines, where limited pilot data are collected before committing to large-scale experiments.

Our key finding is that \textbf{multimodal features enable generalization across parental families, significantly boosting performance under distribution shift.} We see that adding structural and biophysical descriptors to sequence features consistently improves classification. Gains are modest in-distribution (\texttt{scfv\_signature}: +\textbf{3.5} AUROC, +\textbf{3.7} AUPRC) but dramatic under cross-family generalization (\texttt{Parental\_Family}: +\textbf{22.6} AUROC, +\textbf{26.5} AUPRC). Target families show similar boosts (e.g., Fam2: +\textbf{11.2} AUROC, +\textbf{11.2} AUPRC).

Importantly, this result demonstrates that: \textbf{a key factor in bridging the generalization gap is the use of multimodal features}.
Despite the availability of large pretrained encoders, the strongest results come from \emph{simple linear models on well-designed domain-specific features}. Further supporting evidence comes from our ablations (Appendix~\ref{sec:ablation-modality-combination}), which show that the dominant synergy arises between sequence and structure features, while global RMSD alone adds little once explicit structural descriptors are included.

\textbf{Generalization to SEC purity classification task.} Finally, though we focus only on two tasks, synthesis outcomes and yield in the main text, we also evaluated SEC (size-exclusion chromatography) purity prediction as an orthogonal developability assay. As detailed in Appendix~\ref{sec:sec-classification}, our multimodal models achieve strong performance across splits 
further underscoring the broad applicability of our feature representations.

\section{Conclusion and Discussion}
\label{sec:conclusion}

In this work, we studied the problem of antibody reformatting: the problem of converting one antibody format into another to enable its development as a therapeutic.
Specifically, we focused on the the case of reformatting IgG antibodies into single chain variable fragments (scFvs), a key step in unlocking high‐throughput antibody screening and optimization. 

Typically reformatting is a process that involves significant trial-and-error, since a reformatted antibody may synthesize poorly and may have undesirable properties.  Here, we proposed a multimodal machine learning framework, combining sequence, structure, and biophysical representations of antibodies, to be able to predict whether a given reformatting attempt is likely to be successful, and thus reducing the failure rate of reformatting. An important component of our framework is the evaluation methodology, where we  carefully designed evaluation splits that mirror real-world use cases.

In experiments on real-world antibody reformatting data, the  consistent finding is that mulitmodal \textbf{sequence$+$structure$+$biophysical} features together result in the most generalizable models. For protein synthesis failure detection, a model trained on these features achieves an AUROC of over \textbf{88\%} in the most challenging evaluation scenario. Deploying this model could enable prioritization of the most promising candidates during reformatting, resulting in significant cost and time savings. 

A key finding from our experiments is that contrary to recent trends, models trained on embeddings from large, pretrained protein language models (PLMs) do not result in the best performance on this task. Instead, simple domain‐tailored features paired with lightweight predictors perform the best. 
This highlights an important takeaway: for specialized biophysical prediction problems, careful feature design grounded in domain knowledge can outperform large pretrained models, and should remain central in data‐limited settings.

\section*{Acknowledgements}
We would like to thank the BigHat team, especially Ryan Henrici, Emily Delaney, Katrina Stephenson, Duc Huynh, Emily Sever, Anthony Cadena, Noelle Huskey, Taylor Skokan, Nicholas Young, and Lauren Schiff. We also thank the BigHat DS/ML team for productive discussions and insightful suggestions.

\newpage
\bibliographystyle{abbrv}
\bibliography{bib/intro,bib/related_work, bib/multimodal_LLM}

\newpage
\appendix

\newpage

\section{Summary of Notation}
\label{appendix:notation-table}

\begin{table}[h]
\centering
\caption{Extended notation used throughout the paper.}
\label{tab:notation-extended}
\begin{tabular}{ll}
\toprule
\textbf{Symbol} & \textbf{Description} \\
\midrule
$\mathcal{D}$ & Full dataset of $N$ scFv and IgG antibody pairs \\
$(\mathcal{X}_i, \mathbf{y}_i)$ & Input--label pair for the $i$-th scFv and IgG antibody pairs \\
\midrule
$\mathcal{X}$ & Set of input modalities for a construct \\
$\mathbf{x}_{\mathrm{seq}}$ & Sequence features from VH and VL domains \\
$S_{VL}, S_{VH}$ & VL and VH amino acid sequences \\
$L_{VL}, L_{VH}$ & Sequence lengths of VL and VH \\
$\mathbf{E}_{\mathrm{PLM}}(\cdot)$ & Pretrained protein language model encoder \\
$\mathbf{h}_{VL}, \mathbf{h}_{VH}$ & Sequence embeddings for VL and VH \\
\midrule
$\mathbf{x}_{\mathrm{struct}}$ & Structure-derived features \\
$\hat{\mathbf{C}}_{VL}, \hat{\mathbf{C}}_{VH}$ & Predicted C$\alpha$ coordinates for VL and VH \\
$\mathrm{RMSD}_{VL}, \mathrm{RMSD}_{VH}$ & Root mean square deviation per domain \\
\midrule
$\mathbf{x}_{\mathrm{bio}}$ & Biophysical property features derived from CDRs \\
$\mathrm{PSH}$ & Patch Surface Hydrophobicity \\
$\mathrm{PNC}$ & Patch Negative Charge \\
$\mathrm{PPC}$ & Patch Positive Charge \\
$\mathrm{SFvCSP}$ & scFv Charge Separation Product \\
\midrule
$\mathbf{y}_{\mathrm{QC}}$ & Binary Protein Synthesis failure label, $\{0,1\}$ \\
$\mathbf{y}_{\mathrm{yield}}$ & Continuous Protein Synthesis yield label \\
\midrule
$\theta$ & Model parameters \\
$f_\theta(\cdot)$ & Prediction model mapping inputs to $\hat{\mathbf{y}}$ \\
$\hat{\mathbf{y}}$ & Predicted label \\
$\mathcal{L}(\cdot, \cdot)$ & Task-appropriate loss function \\
\midrule
$\mathcal{D}_{\mathrm{train}}, \mathcal{D}_{\mathrm{val}}, \mathcal{D}_{\mathrm{test}}$ & Train/validation/test subsets \\
\textsc{scFv} split & Random partition over unique scFv signatures \\
\textsc{Target-family} split & Hold-out of target family with few-shot fine-tuning \\
\textsc{Parental-family} split & Zero-shot hold-out of parental family \\
\bottomrule
\end{tabular}
\end{table}

\section{Dataset Statistics}
\label{appendix:dataset-statistics}

\subsection*{Dataset Overview}
\begin{figure}[!t]
    \vspace{-3.0mm}
    \centering
    \includegraphics[width=\textwidth]{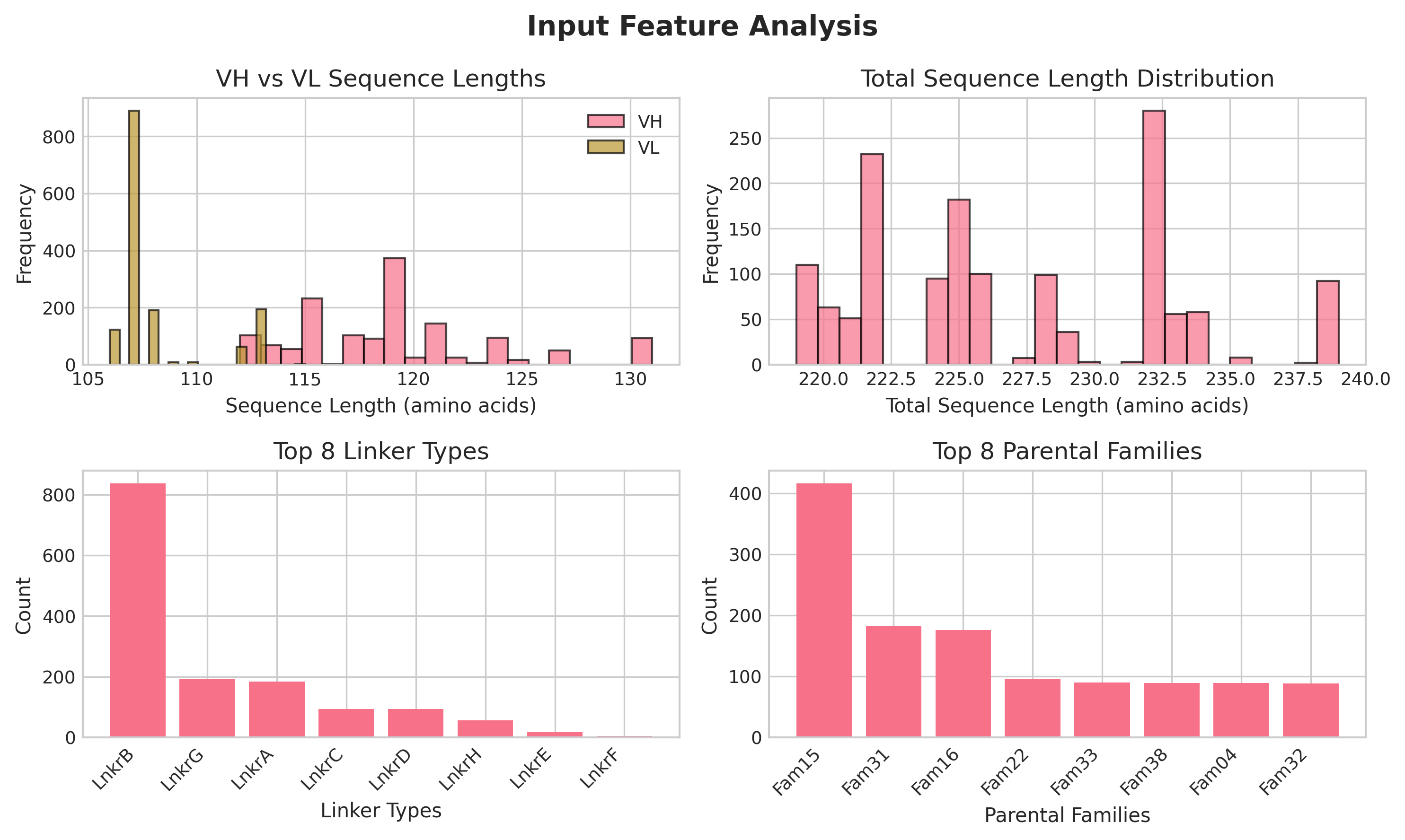}
    \caption{Exploratory data analysis of input features.}
    \label{fig:appendix-eda-input-feats}
\end{figure}

Our scFv$\rightarrow$IgG Reformatting Dataset comprises 1{,}477 unique scFv signatures, spanning 52 parental families.
Figure \ref{fig:appendix-eda-input-feats} shows exploratory data analysis of input features.

\subsection*{Input Features}
Sequence length is consistent over scFv$\rightarrow$IgG pairs. VH domains average $118.9 \pm 4.8$ amino acids, VL domains average $108.1 \pm 2.2$ amino acids, and the combined sequence length averages $227.0 \pm 5.6$ amino acids. The dataset includes 8 linker types, 2 domain orderings, and covers 52 distinct parental families, providing diversity across scFv construct designs while maintaining tight length distributions.

\subsection*{Target Variables}
The primary regression target, yield, has a mean of $31.33 \pm 51.44~\mathrm{ng}/\mu\mathrm{L}$ with a wide dynamic range from $3.50$ to $513.58~\mathrm{ng}/\mu\mathrm{L}$. For classification tasks, 40.9\% of constructs failed protein synthesis,

\subsection*{ML Considerations}
With $n=1{,}477$ observations, the dataset is limited for modern machine learning and would benefit from additional data for higher-capacity models. The protein synthesis failure outcome exhibits moderate class imbalance (40.9\% failure vs.\ 59.1\% pass), which is addressed with appropriate metrics (AUROC and AUPRC).
The broad yield range is suitable for regression, while the rich sequence-based inputs (VH and VL) provide strong signal for representation learning and featurization.

\vspace{-0.4em}
\section{Details of Method Section}
\vspace{-0.2em}
\label{appendix:method-details}

\subsection{Problem Formulation and Notation}
\label{sec:problem-notation}

Let each antibody be represented by a set of input modalities \(\mathcal{X} = \{\mathbf{x}_{\mathrm{seq}}, \mathbf{x}_{\mathrm{struct}}, \mathbf{x}_{\mathrm{bio}}\}\), where \(\mathbf{x}_{\mathrm{seq}}\) encodes the VH and VL amino acid sequences (over the 20-amino acid alphabet), \(\mathbf{x}_{\mathrm{struct}}\) encodes structure-derived descriptors from predicted 3D structures, and \(\mathbf{x}_{\mathrm{bio}}\) contains biophysical properties. The target variable \(\mathbf{y}\) corresponds either to a protein synthesis success or failure, \(\mathbf{y} \in \{0,1\}\), or a continuous synthesis yield value, \(\mathbf{y} \in \mathbb{R}\) (where higher is better). Given a dataset \(\mathcal{D} = \{(\mathcal{X}_i, \mathbf{y}_i)\}_{i=1}^N\), the objective is to learn a mapping \(f_\theta: \mathcal{X} \mapsto \hat{\mathbf{y}}\) that minimizes a task-appropriate loss \(\mathcal{L}(\mathbf{y}, \hat{\mathbf{y}})\) and generalizes to unseen antibody families, 
i.e., correctly predicting synthesis success for novel families not observed during training. 

\subsection{Dataset Construction}
\label{sec:dataset}

\noindent
\textbf{Inputs and targets.}  
Our dataset consists of IgG$\rightarrow$scFv reformatting experiments, where full-length immunoglobulin G (IgG) antibodies are converted into compact single-chain variable fragments (scFvs).  These experiments were conducted across multiple antibody optimization campaigns.  
Each scFv is represented by: (i) VH and VL amino acid sequences, (ii) the linker sequence connecting the domains, (iii) the domain ordering (VH--VL or VL--VH), and (iv) the parental family identifier from which the VH and VL are derived.  
We train separate models for two primary tasks: (1) \emph{protein synthesis outcome classification}, where the target variable $y_{\mathrm{QC}} \in \{0,1\}$ indicates whether a reformatted scFv has synthesized adequately or not, and (2) \emph{yield regression}, where $y_{\mathrm{yield}} \in \mathbb{R}$ measures synthesis yield in ng/$\mu$L.

\noindent
\textbf{scFv signature and aggregation.}  
We define the \emph{scFv signature} of an input scFv as the tuple:
\[
\textsc{Sig} = (\mathrm{VH}, \mathrm{VL}, \mathrm{linker}, \mathrm{orientation}),
\]
where $\mathrm{VH}$ and $\mathrm{VL}$ are the amino acid sequences of the heavy and light chain variable regions, $\mathrm{linker}$ is the connecting peptide, and $\mathrm{orientation}$ specifies the domain order.  
scFvs sharing the same signature are considered equivalent: their target values are averaged.  
After aggregation, the dataset contains $N=1{,}477$ unique scFv signatures drawn from $56$ parental families across $7$ independent antibody optimization campaigns. Summary statistics are provided in Appendix~\ref{appendix:dataset-statistics}.

\noindent
\textbf{Parental family generalization.}  Within a parental family, scFv sequences are often highly similar, differing by only a few mutations. However, across families, divergence is much greater, creating a strong distribution shift between families. This makes generalization to unseen families a central challenge for our models. This motivates our evaluation strategy, discussed next.

\subsection{Evaluation Protocol}
\label{appendix:evaluation}

\begin{wrapfigure}{r}{0.64\textwidth}
\centering
\vspace{-10mm}
\includegraphics[width=0.99\linewidth]{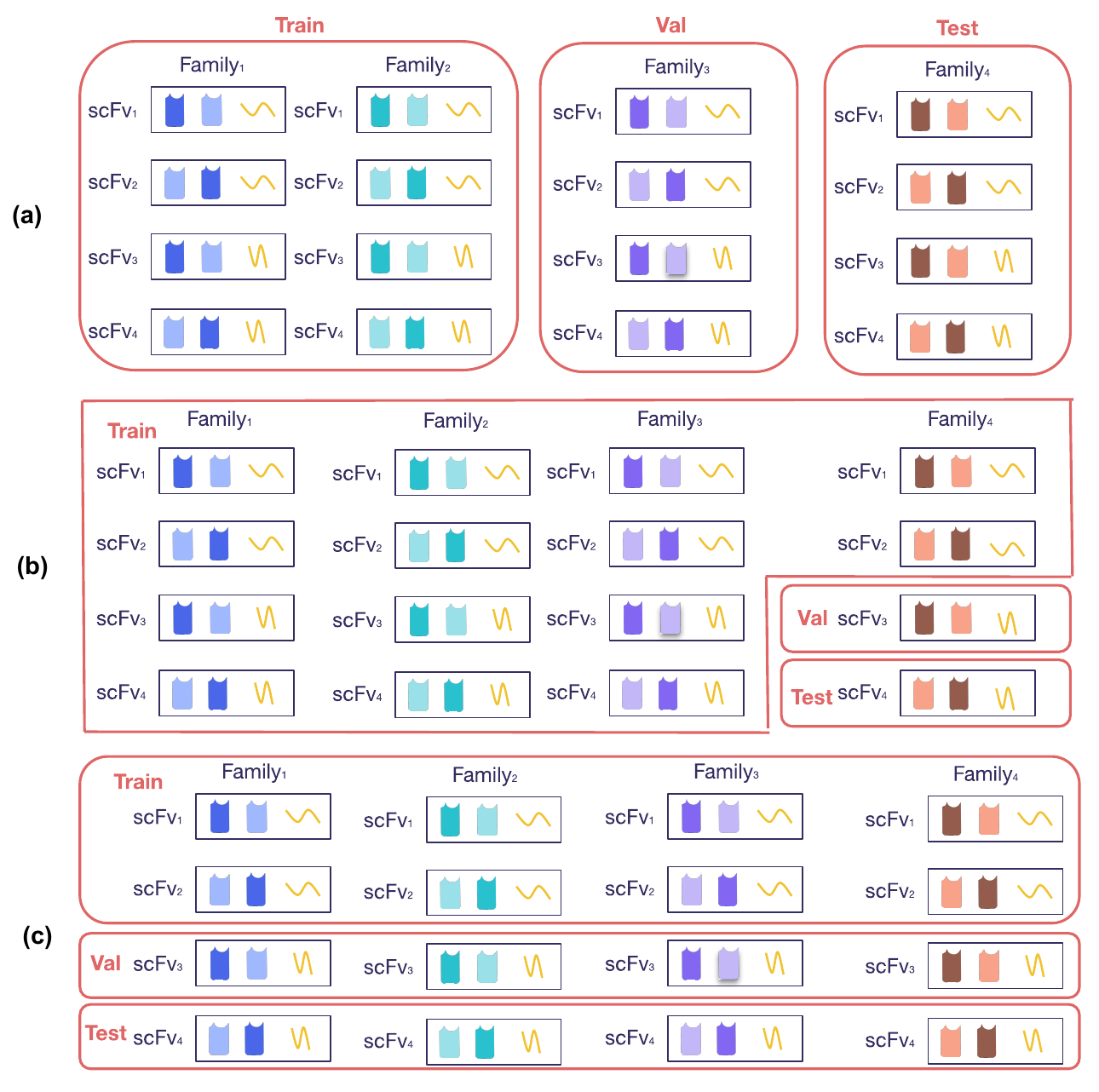}
\vspace{-1mm}
\caption{Three data splits illustrated. (a) Parental Family split. (b) Target Family split. (c) scFv split.}
\label{fig:method-three-splits}
\vspace{-2mm}
\end{wrapfigure}

We evaluate model generalization under three disjoint data partitioning schemes, each corresponding to a realistic deployment scenario (Figure \ref{fig:method-three-splits}): (1) \emph{Parental Family split} (\textsc{New Parental, No Data}): entire parental families are held out from training, forcing zero-shot prediction on novel families with no prior \textit{in vitro} data. (2) \emph{Target Family split} (\textsc{One Batch from New Antibody}): a small batch from the target family is included in training along with data from other families; the remaining scFvs from the target family are used for validation and testing. (3) \emph{scFv split} (\textsc{Lots of Data for All Families}): all families appear in all splits, but individual scFv signatures are unique to a single split, preventing data leakage while maximizing same-family context.

\subsection{Feature Representations}
\label{sec:features}

Our modeling approach integrates three complementary feature modalities: sequence, structure, and biophysical properties. Each modality captures distinct aspects of the scFv--parental IgG relationship, and all features are precomputed and frozen prior to model training.

\paragraph{Sequence-based features.}  
VH and VL amino acid sequences are AHo-aligned\citep{honegger2001aho} to a fixed length ($L_{\mathrm{VL}}=152$, $L_{\mathrm{VH}}=152$) and one-hot encoded. Domain orientation (VH--VL or VL--VH) and linker peptide type are represented as categorical one-hot variables and concatenated with the sequence encodings.  
Given the potential of pretrained sequence models to encode rich information about an input protein, we also evaluate predicting reformatting success using frozen embeddings from two pLMs. First, we consider AbLang~\citep{ablang2_bioinf2024} (heavy/light encoders separately), which is an antibody-specific pLM trained on a large dataset of human antibody sequences. Secondly, we consider ISM~\citep{dreyer2023inverse}, a fine-tuned ESM \citep{rives2021biological} model that is trained to encode structural information in addition to sequence. For both models, residue-level embeddings are computed once and then mean-pooled across the sequence to yield a fixed-length representation. These are kept fixed during downstream training. Our intuition is that such pretrained encoders capture general antibody sequence statistics (e.g., conserved CDR motifs, germline variation) that may aid generalization across families.

\paragraph{Structure-based features.}  
For each unique scFv and its corresponding parental IgG, we predict full-atom 3D structures using Boltz-2~\citep{passaro2025boltz}, a structure prediction model with accuracy comparable to AlphaFold3 across diverse proteins. After performing rigid body alignment between an scFv/IgG pair we compute descriptors from these predictions: (i) the global RMSD of C$\alpha$ atoms between VH/VL domains of the parental IgG and its reformatted scFv, and (ii) per-residue concatenation of aligned parental and scFv C$\alpha$ coordinates with gap indicators, preserving spatial correspondence. 

The motivation is grounded in structural biology, since the function of a protein is intimately related to function, alterations of the structure of individual domains between formats may be related to reformatting success. RMSD serves as a coarse global descriptor (A detailed analysis of RMSD distributions across families is provided in Appendix~\ref{appendix:analysis-structure-per-family}). The per-residue coordinate features offer a more fine-grained view, potentially capturing subtle local rearrangements that influence reformatting outcomes. Representative overlays of VH and VL domains are shown in Figures~\ref{fig:vh_overlay} and~\ref{fig:vl_overlay}.

To benchmark these domain-inspired descriptors against widely used pretrained structural models, we also extract structure-derived embeddings using two models: (i) AbMPNN \citep{dreyer2023inverse} (an antibody inverse folding model that encodes sequences in the context of 3D backbones) and (ii) DPLM2 \citep{wangdplm} (a structure-augmented protein language model). Both provide residue-level embeddings that are mean-pooled to fixed-length vectors for downstream tasks. Our rationale is that inverse folding and structure-augmented pLMs may capture geometric constraints and stability signals that purely sequence-based encoders miss, and thus could help with predicting reformatting success.

\begin{figure*}[t]
  \centering
  \begin{minipage}[t]{0.45\textwidth}
    \centering
    \includegraphics[width=\linewidth]{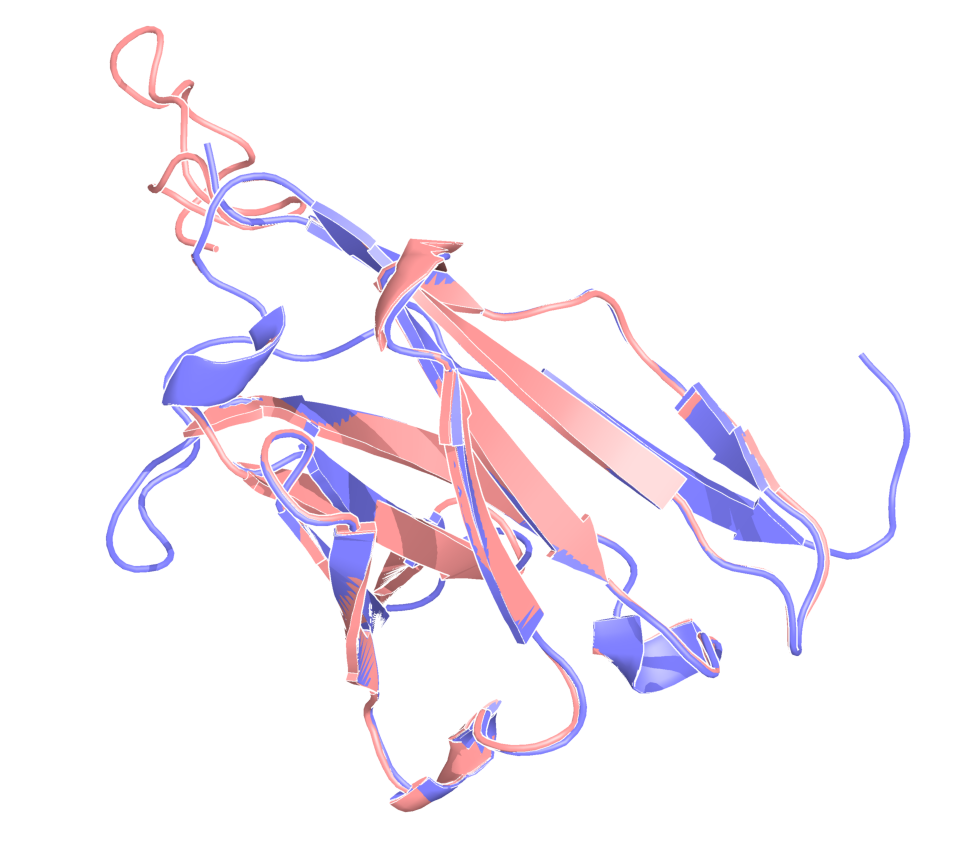}
    \caption{Overlay of predicted VH domain structure between the starting IgG (Purple) and reformatted to scFv (pink). In this example Boltz-2 predicts significant structural alteration upon reformatting.}
    \label{fig:vh_overlay}
  \end{minipage}\hfill
  \begin{minipage}[t]{0.45\textwidth}
    \centering
    \includegraphics[width=\linewidth]{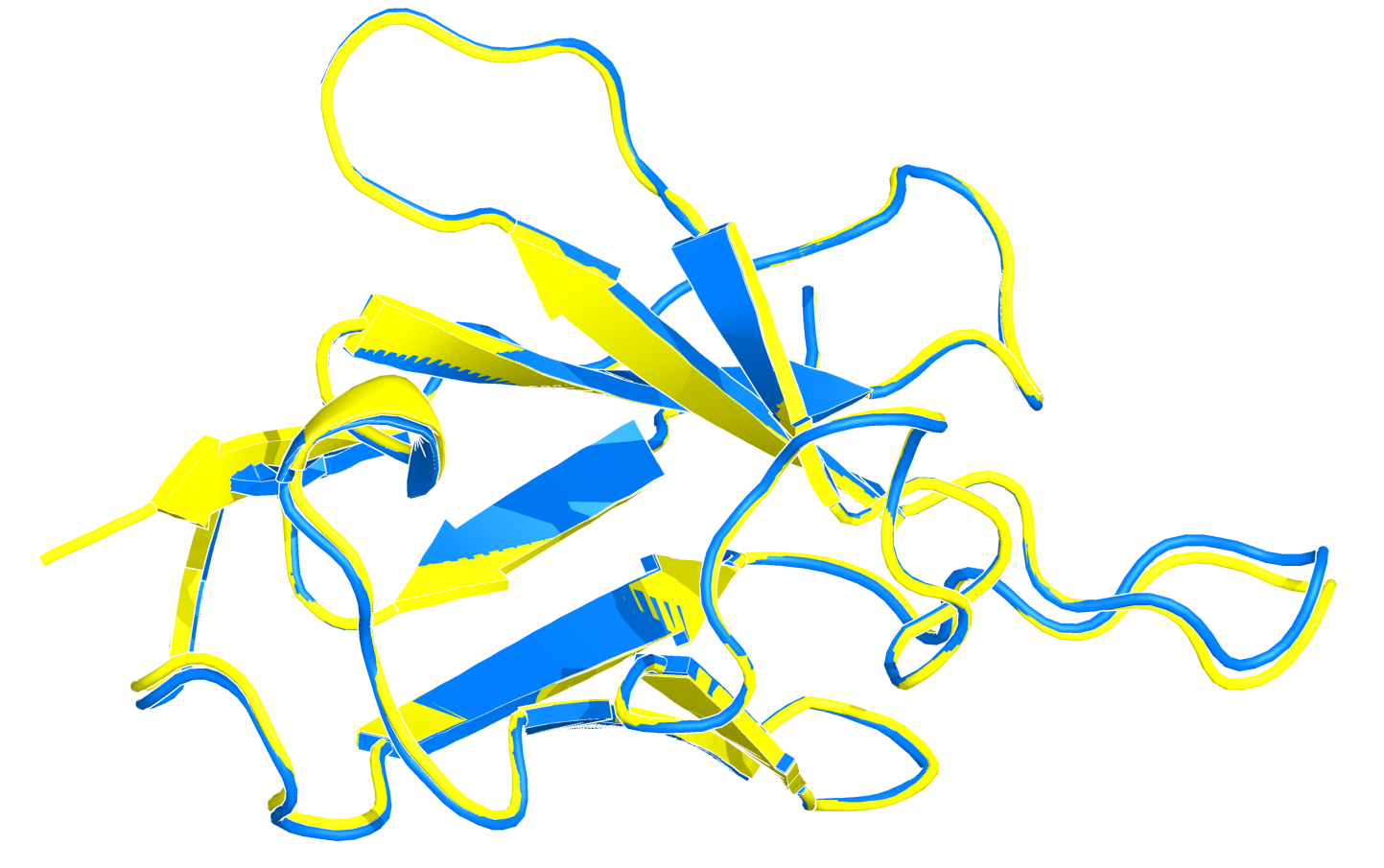}
    \caption{Overlay of predicted VL domain structure between the starting IgG (blue) and reformatted to scFv (yellow).}
    \label{fig:vl_overlay}
  \end{minipage}
\end{figure*}

\paragraph{Biophysical features.}  
From predicted scFv structures, we compute developability metrics using the NaturalAntibody \citep{naturalantibody} platform. Key features derived from the CDR regions include Patch Surface Hydrophobicity (PSH), Patch Negative Charge (PNC), Patch Positive Charge (PPC), and scFv Charge Separation Product (SFvCSP). These are metrics which could be expected to be associated with general antibody stability and expressability. If a score cannot be computed due to modeling failure, the dataset mean is imputed.

\subsection{Model Architectures}
\label{sec:architectures}

\paragraph{Baseline models.}  
We evaluate linear baselines for both protein synthesis outcome classification and yield regression. For classification, we use logistic regression with either L1 or L2 regularization (the choice of regularization is treated as one hyperparameter during tuning); for regression, we use ordinary least squares with optional regularization. Two input configurations are compared: (i) one-hot AHo-aligned VH and VL sequences concatenated with one-hot domain orientation and linker encodings, and (ii) frozen pLM embeddings. All linear models are implemented in \texttt{scikit-learn} and trained with default solvers.

\paragraph{Embedding-based models.}  
We evaluate a fixed-architecture multilayer perceptron (MLP) on frozen pLM embeddings (which outperformed linear models on pLM embeddings). VH and VL embeddings are concatenated prior to the MLP, which contains two hidden layers with ReLU activations and dropout. Hyperparameters such as dropout and learning rate are tuned via grid search, while architecture depth and width remain fixed. Models are trained with AdamW and early stopping.

\paragraph{Multimodal ML models.}  

To assess complementarity between modalities, we train simple linear models using combinations of sequence, structure, and biophysical features. Hyperparameter tuning follows the same protocol as for the embedding-based models.

\section{Hyperparameter Details}
\label{appendix:detail-hyperparameter}

We summarize the hyperparameters used for all models evaluated in this work. Unless otherwise noted, all models were trained with the AdamW optimizer and early stopping on the validation loss. Hyperparameters were selected via grid or Optuna \citep{akiba2019optuna} search on the validation set. Below we report both the search ranges and the final values used in our experiments.

\paragraph{Linear models.}  
\textit{Search space:} regularization strength $C \in \{0.01, 0.1, 1, 10\}$, penalty $\in \{\ell_1, \ell_2\}$.  
\textit{Final choice:} for linear regression, $\ell_2$ penalty with $C=0.01$; for logistic regression, $\ell_2$ penalty with $C=10$.  

\paragraph{Pretrained embeddings + MLP.}  
\textit{Search space:} hidden dimension $\in \{64, 128,256\}$, dropout $\in \{0.1, 0.2,0.3\}$, learning rate $\in \{10^{-3}, 10^{-4}\}$, batch size $\in \{32, 64\}$, and a binary flag for using a linear head.  
\textit{Final choice:} a two-layer MLP with hidden dimension $128$ or $256$, ReLU activations, dropout $0.2$, learning rate $1\times10^{-4}$, and batch size $32$ or $64$.  

\paragraph{1D CNNs.}  
\textit{Search space (Optuna):} number of dilated convolutional layers $[1,5]$, expansion factor $[1.0, 4.0]$, representation dimension $\{16,32,64,128\}$, batch normalization $\in \{\texttt{true}, \texttt{false}\}$, learning rate $[10^{-4}, 10^{-2}]$ (log-uniform), batch size $\{16,32,64\}$, and training epochs $[10,50]$.  
\textit{Final choice:} for classification, $5$ dilated conv layers, representation dimension $32$, expansion factor $1.4$, no batch normalization, learning rate $3.8\times10^{-4}$, batch size $32$, and training for $13$ epochs. For regression, $1$ dilated conv layer, representation dimension $16$, expansion factor $2.7$, no batch normalization, learning rate $1.8\times10^{-4}$, batch size $32$, and training for $10$ epochs.

\section{Ablation Studies and In-depth Analysis}
\label{sec:ablation-analysis}

\subsection{Ablation studies of different modality combinations}
\label{sec:ablation-modality-combination}

\begin{figure}[!t]
\vspace{-3.0mm}
    \centering 
    \includegraphics[width=\textwidth]{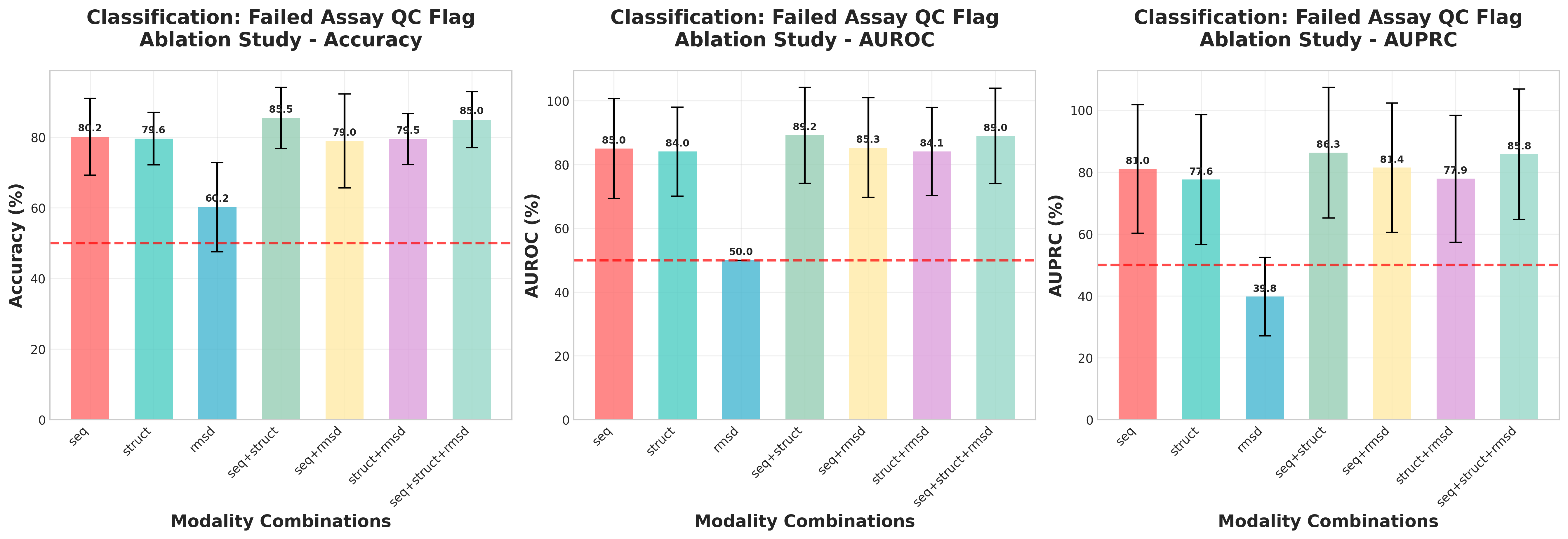}
    \caption{Ablation study on different modality combination on QC classification task.}
    \label{fig:ablation-modality-combination}
\end{figure}

We ablate seven modality configurations on the \emph{Failed Assay QC} classifier—\texttt{seq}, \texttt{struct}, \texttt{rmsd}, all pairwise combinations, and \texttt{seq+struct+rmsd}. Figure~\ref{fig:ablation-modality-combination} summarizes Accuracy, AUROC, and AUPRC (mean with standard-deviation bars; red dashed line denotes the random baseline).

Across all metrics, \textbf{\texttt{seq+struct}} is uniformly strongest, reaching \textbf{85.5 Acc}, \textbf{89.2 AUROC}, and \textbf{86.3 AUPRC}. This reflects clear complementarity: relative to the best single-modality baselines, \texttt{seq+struct} gains $+5.3$/$+4.2$/$+5.3$ points over \texttt{seq} and $+5.9$/$+5.2$/$+8.7$ over \texttt{struct} on Accuracy/AUROC/AUPRC, respectively. In contrast, \texttt{rmsd} alone carries little predictive signal (AUROC $\approx\!50.0$, AUPRC $39.8$), and adding it to either \texttt{seq} or \texttt{struct} yields at most marginal changes (\texttt{seq+rmsd}: 85.3 AUROC, 81.4 AUPRC; \texttt{struct+rmsd}: 84.1 AUROC, 77.9 AUPRC). Notably, the full tri-modal model \texttt{seq+struct+rmsd} does not improve over \texttt{seq+struct} (89.0 vs.\ 89.2 AUROC; 85.8 vs.\ 86.3 AUPRC), suggesting that global RMSD largely overlaps with information already captured by explicit structural descriptors and can introduce redundant or noisy signal.

\textbf{Takeaway.} The dominant synergy is between sequence and structure: residue-level biochemical constraints from \texttt{seq} and geometric compatibility from \texttt{struct} combine to drive the best generalization, while RMSD—being a coarse, global deviation measure—adds little once structural features are explicitly modeled.

\subsection{Different structure–feature distributions across parental families}
\label{appendix:analysis-structure-per-family}

\begin{figure}[H]
    \vspace{-3.0mm}
    \centering
    \includegraphics[width=\textwidth]{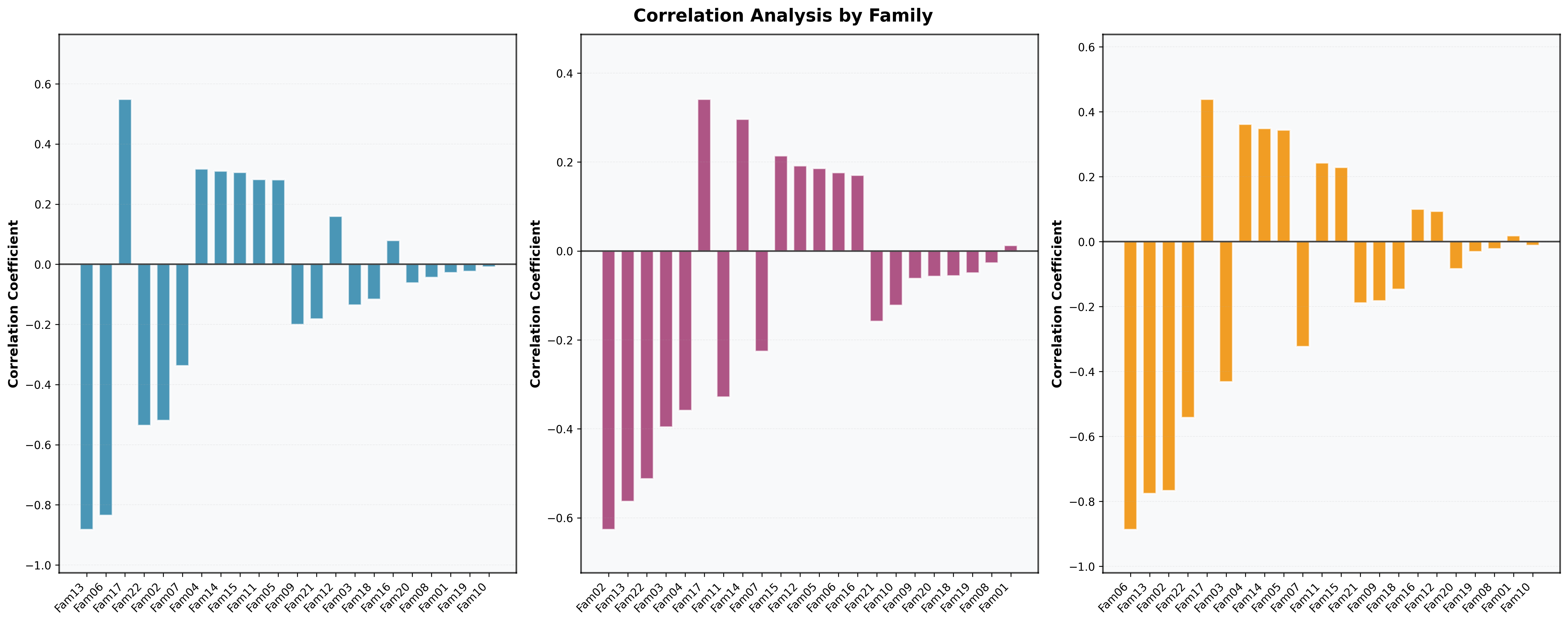}
    \caption{\textbf{Family-specific RMSD–yield relationships.} Pearson correlations between yield and VH RMSD (left), VL RMSD (middle), and VH+VL RMSD (right), computed \emph{within} each parental family (families sorted by correlation magnitude). The wide dispersion and sign changes indicate heterogeneous, family-dependent structure–yield trends.}
    \label{fig:rmsd-correlation-by-family}
\end{figure}

To probe why RMSD features offer inconsistent gains, we quantify the within–parental-family association between structural deviation and protein yield. For each family, we compute the Pearson correlation between yield and (i) VH RMSD, (ii) VL RMSD, and (iii) their sum. Figure~\ref{fig:rmsd-correlation-by-family} reveals pronounced heterogeneity: several families exhibit a \emph{positive} RMSD–yield relationship (larger deviations correlate with higher yield), whereas others show the opposite trend, and many lie near zero. These sign flips persist across VH, VL, and combined RMSD, and are not explained by sample size alone (family-wise $n$ is indicated atop each bar). 

This family-specific behavior implies non-stationarity in the mapping from global structural deviation to expression outcome. A model trained on pooled data with only global structure descriptors faces a conflicting supervision signal and will tend to regress toward a weak average effect, limiting the utility of RMSD as a standalone predictor and explaining its marginal contribution once richer structural descriptors are already present. The absence of explicit parental-family information further prevents the learner from capturing divergent RMSD–yield regimes.

\textbf{Implication.} Incorporating family context—e.g., parental-family identifiers, sequence backbones, or a conditional/mixture-of-experts gate—should allow the model to adapt its structural-to-yield mapping across families. This aligns with our ablation in Section \ref{sec:ablation-modality-combination}: the strongest performance arises when sequence (a proxy for family identity and local biophysics) is fused with structure, while global RMSD alone is insufficient.

\subsection{Case study of the \textit{Fam1} parental family}
\label{sec:case-study-tinurilimab}

We analyze the model’s behavior on a single parental family (\textit{Fam1}) to understand operational impact on protein synthesis failure screening. On the held-out test set of $55$ variants, the classifier attains \textbf{Recall $=100\%$} (no good scFv missed), \textbf{Precision $=87.2\%$} (five bad scFvs flagged as “pass”), and \textbf{Accuracy $=90.9\%$}. In other words, the model screens out almost all low-quality candidates while retaining every high-quality one, eliminating wasted experiments from false negatives and confining errors to a small number of false positives.

From an operational perspective, we evaluate an actionable \emph{screen-then-confirm} policy that only advances candidates predicted to successful protein synthesis. Under this policy, the fraction of executed experiments that yield a true pass equals the classifier’s positive predictive value,
\[
\text{Efficiency} \;=\; \frac{\mathrm{TP}}{\mathrm{TP}+\mathrm{FP}} \;=\; \text{Precision}.
\]
Relative to a trial-and-error baseline efficiency of \textbf{61.8\%}, our multimodal model would raise efficiency to \textbf{87.2\%}, an \textbf{absolute gain of $+25.4$ points} and a \textbf{$1.4\times$} multiplicative improvement. Practically, this means more assays are spent confirming genuinely promising variants rather than testing poor candidates.

\textbf{Takeaway.} Within this family, the model provides a high-recall triage mechanism—\emph{no good scFv is missed}—while substantially improving experimental efficiency. We note, however, that this is a single-family case study and results can vary with stochastic training and family-specific covariate shift; broader prospective evaluation and calibration across families remain important.

\subsection{SEC classification}
\label{sec:sec-classification}

\begin{figure}[!t]
\caption{Linear model performance on multimodal feature input for SEC purity ($y=\mathds{1}[\text{\% area under the main peak 280nm}\ge 90\%]$).}
    \vspace{-3.0mm}
    \centering
    \includegraphics[width=.6\textwidth]{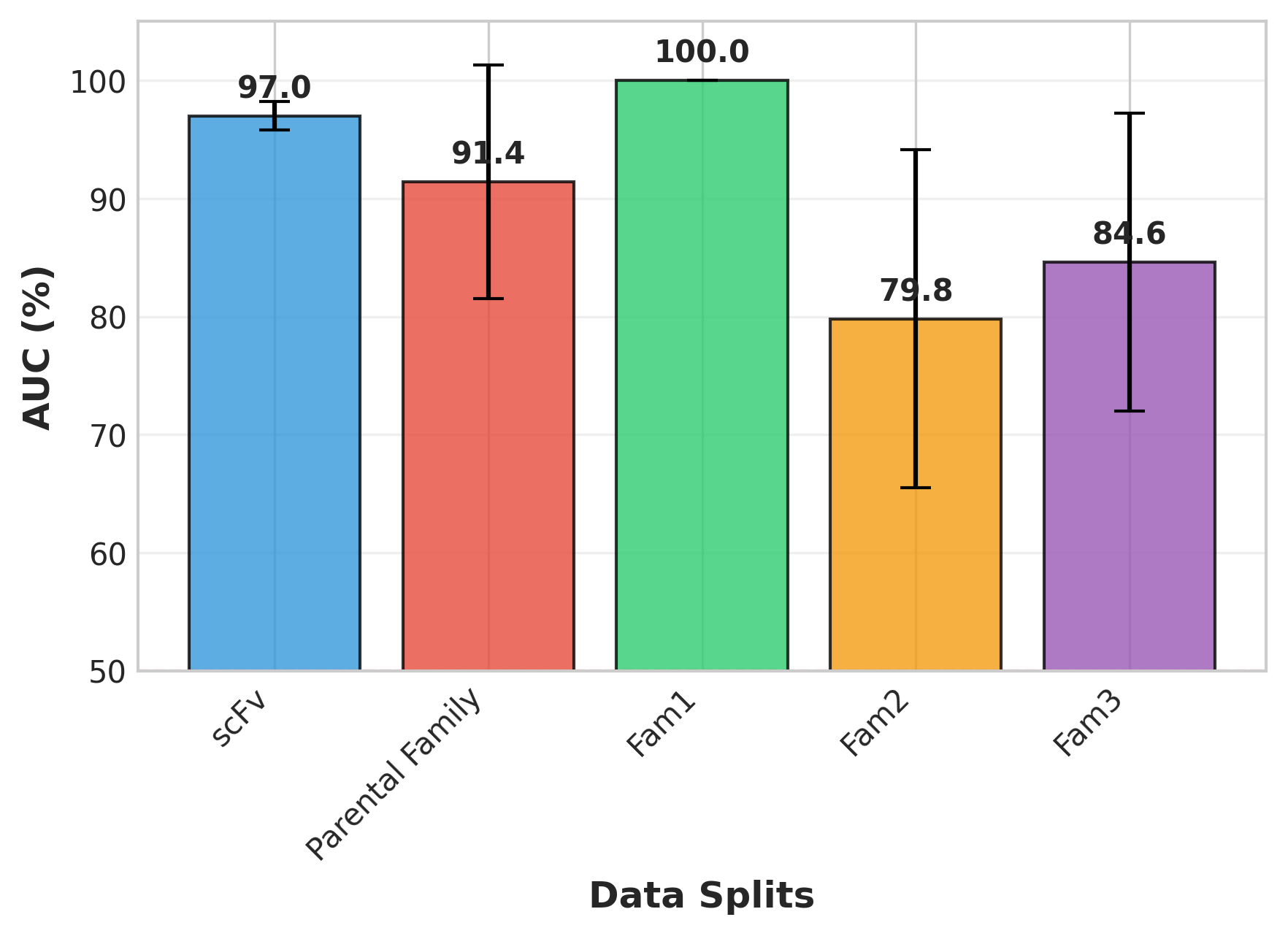}

   \label{fig:analysis-sec-linear}
\end{figure}

We formulate SEC (size‐exclusion chromatography) purity prediction as a binary classification task with label
$\;y=\mathbf{1}[\text{\% area under main peak 280 nm}\ge 90\%]\,$ and evaluate generalization across several realistic data partitions that vary in sequence and family composition. The model attains strong and stable performance across splits, with mean AUCs of \textbf{97.0} on the \texttt{scFv} split, \textbf{91.4} on the \texttt{Parental Family} split, and a perfect \textbf{100.0} on \texttt{Fam1}. Performance is lower but still competitive for more distributionally shifted families, achieving \textbf{79.8} on \texttt{Fam2} and \textbf{84.6} on \texttt{Fam\_102}. Error bars (standard deviation across runs) are narrow on \texttt{scFv} and \texttt{Fam1}, but widen on family‐held‐out splits, indicating increased variance when the model is asked to extrapolate to unseen parental backbones. Taken together, these results suggest that the learned representation captures robust determinants of SEC purity and transfers well, with degradation primarily attributable to family‐specific covariate shift.

\section{Extended Experiment Results for Protein Synthesis Failure Classification}

We present the accuracy metrics for protein synthesis failure classification in the following tables. Table \ref{tab:acc-seq-classification} shows the accuracy for linear model compare to PLM embeddings with sequence features only. Table \ref{tab:acc-struct-classification} shows the accuracy for structure input with Logistic Regression and 1D-CNN. Table \ref{tab:acc-multimodal-classification} shows the accuracy for multimodal input.

\begin{table*}[!t]
\centering
\caption{Classification (protein synthesis failure) with sequence and/or structural features. We compare PLM embeddings with an MLP (\texttt{AbLang+MLP}, \texttt{ISM+MLP}), a structure GNN with an MLP (\texttt{AbMPNN+MLP}), a structure-augmented PLM with an MLP (\texttt{DPLM2+MLP}), and a one-hot logistic regression baseline (\texttt{LogisticReg}). Best results per column are \underline{\textbf{underlined in bold}}.}
\label{tab:acc-seq-classification}
\resizebox{.8\textwidth}{!}{%
\begin{tabular}{l|l|c|c}
\toprule
\textbf{Model (Acc. in \%)} & \textbf{Features} &
\textbf{scfv\_signature split} &
\textbf{Parental\_Family split} \\
\midrule
AbLang+MLP   & \texttt{vhvl\_only}    & 80.00\scriptsize{$\pm$1.83} & 59.64\scriptsize{$\pm$16.89} \\
ISM+MLP      & \texttt{vhvl\_only}    & 71.42\scriptsize{$\pm$1.85} & 56.64\scriptsize{$\pm$12.07} \\
DPLM2+MLP    & \texttt{vhvl+struct}   & 72.84\scriptsize{$\pm$2.06} & 48.67\scriptsize{$\pm$9.02}  \\
AbMPNN+MLP   & \texttt{struct\_only}  & 60.47\scriptsize{$\pm$1.23} & 49.85\scriptsize{$\pm$7.98}  \\
\rowcolor{LightCyan}
LogisticReg  & \texttt{vhvl\_only}    & \underline{\textbf{83.58}}\scriptsize{$\pm$1.69} & \underline{\textbf{60.98}}\scriptsize{$\pm$12.53} \\
\bottomrule
\end{tabular}}
\end{table*}

\begin{table*}[!t]
\centering
\small
\setlength{\tabcolsep}{4pt} 
\renewcommand{\arraystretch}{1.1} 
\caption{Protein synthesis failure classification using per-residue 3D coordinate features (\texttt{3D\_coord}). Rows correspond to data splits; columns compare \texttt{LogisticReg} vs.\ \texttt{1DCNN} head-to-head. Values are mean$\pm$std across runs. Best per split is \underline{\textbf{underlined in bold}}.}
\label{tab:acc-struct-classification}
\begin{tabular}{l
                c c
                c c
                c c}
\toprule
\multirow{2}{*}{\textbf{Split}} 
 & \multicolumn{2}{c}{\textbf{Accuracy}} 
 & \multicolumn{2}{c}{\textbf{AUROC}} 
 & \multicolumn{2}{c}{\textbf{AUPRC}} \\
 & \textbf{LogisticReg} & \textbf{1DCNN} 
 & \textbf{LogisticReg} & \textbf{1DCNN} 
 & \textbf{LogisticReg} & \textbf{1DCNN} \\
\midrule
 \texttt{scfv\_signature} 
 & 72.00{\scriptsize$\pm$2.00} & \underline{\textbf{76.00}}{\scriptsize$\pm$2.00}
 & 77.00{\scriptsize$\pm$1.00} & \underline{\textbf{82.00}}{\scriptsize$\pm$2.00}
 & 71.00{\scriptsize$\pm$2.00} & \underline{\textbf{79.00}}{\scriptsize$\pm$2.00} \\
\texttt{Parental\_Family} 
 & 52.00{\scriptsize$\pm$9.00} & \underline{\textbf{54.00}}{\scriptsize$\pm$12.00}
 & 52.00{\scriptsize$\pm$12.00} & \underline{\textbf{57.00}}{\scriptsize$\pm$13.00}
 & 48.00{\scriptsize$\pm$8.00} & \underline{\textbf{52.00}}{\scriptsize$\pm$13.00} \\
\texttt{Fam1} 
 & \underline{\textbf{56.00}}{\scriptsize$\pm$5.00} & 46.00{\scriptsize$\pm$13.00}
 & \underline{\textbf{58.00}}{\scriptsize$\pm$6.00} & 49.00{\scriptsize$\pm$22.00}
 & \underline{\textbf{70.00}}{\scriptsize$\pm$5.00} & 64.00{\scriptsize$\pm$15.00} \\

\texttt{Fam2} 
 & 70.00{\scriptsize$\pm$3.00} & \underline{\textbf{74.00}}{\scriptsize$\pm$0.00}
 & \underline{\textbf{59.00}}{\scriptsize$\pm$8.00} & 49.00{\scriptsize$\pm$7.00}
 & \underline{\textbf{39.00}}{\scriptsize$\pm$6.00} & 28.00{\scriptsize$\pm$2.00} \\
\texttt{Fam3} 
 & \underline{\textbf{79.00}}{\scriptsize$\pm$5.00} & 38.00{\scriptsize$\pm$8.00}
 & \underline{\textbf{79.00}}{\scriptsize$\pm$8.00} & 73.00{\scriptsize$\pm$10.00}
 & \underline{\textbf{90.00}}{\scriptsize$\pm$5.00} & 89.00{\scriptsize$\pm$4.00} \\

\bottomrule
\end{tabular}
\end{table*}

\begin{table*}[!t]
\centering
\small
\setlength{\tabcolsep}{4pt} 
\renewcommand{\arraystretch}{1.1} 
\caption{Protein synthesis failure classification using \emph{sequence+structure+biophysics} features (\texttt{multimodal}) vs.\ sequence-only (\texttt{seq\_only}) with a linear classifier. Rows correspond to data splits; values are mean$\pm$std across runs. Best per split is \underline{\textbf{underlined in bold}}.}
\label{tab:acc-multimodal-classification}
\begin{tabular}{l c c}
\toprule
\textbf{Split (Acc. in \%)} 
 & \texttt{multimodal} & \texttt{seq\_only} \\
\midrule
\texttt{scfv\_signature} 
 & \underline{\textbf{85.24}}{\scriptsize$\pm$4.33} & 83.58{\scriptsize$\pm$1.69} \\
\texttt{Parental\_Family} 
 & \underline{\textbf{85.67}}{\scriptsize$\pm$7.60} & 60.98{\scriptsize$\pm$12.53} \\
\texttt{Fam1} 
 & \underline{\textbf{86.55}}{\scriptsize$\pm$4.61} & 80.18{\scriptsize$\pm$4.91} \\
\texttt{Fam2} 
 & \underline{\textbf{77.78}}{\scriptsize$\pm$4.76} & 75.19{\scriptsize$\pm$3.43} \\
\texttt{Fam3} 
 & \underline{\textbf{84.81}}{\scriptsize$\pm$10.79} & 79.26{\scriptsize$\pm$3.78} \\
\bottomrule
\end{tabular}
\end{table*}

\section{Extended Experiment Results for Protein Synthesis Yield Regression}

We present the Pearson correlation and Spearman correlation metrics for protein synthesis yield regression in the following tables. Table \ref{tab:appendix-seq-regression} shows the Pearson correlation and Spearman correlation for linear model compare to PLM embeddings with sequence features only. Table \ref{tab:appendix-struct-regression} shows the Pearson correlation and Spearman correlation for structure input with Logistic Regression and 1D-CNN.

\begin{table*}[!t]
\centering
\caption{Regression (protein yield) with sequence and/or structural features. We report mean$\pm$std across runs. Best results per column are \underline{\textbf{underlined in bold}}.}
\label{tab:appendix-seq-regression}
\resizebox{0.8\textwidth}{!}{%
\begin{tabular}{l|l|cc|cc}
\toprule
\textbf{Model} & \textbf{Features} &
\multicolumn{2}{c|}{\textbf{scfv\_signature split}} &
\multicolumn{2}{c}{\textbf{Parental\_Family split}} \\
\cmidrule(lr){3-4} \cmidrule(lr){5-6}
& & \textbf{Pearson} & \textbf{Spearman} & \textbf{Pearson} & \textbf{Spearman} \\
\midrule
AbLang+MLP   & \texttt{vhvl\_only}    & \underline{\textbf{0.562}}\scriptsize{$\pm$0.018} & 0.610\scriptsize{$\pm$0.012}  & 0.014\scriptsize{$\pm$0.161}  & 0.052\scriptsize{$\pm$0.331} \\
ISM+MLP      & \texttt{vhvl\_only}    & 0.173\scriptsize{$\pm$0.019}  & 0.368\scriptsize{$\pm$0.037}  & -0.110\scriptsize{$\pm$0.142} & -0.104\scriptsize{$\pm$0.178} \\
DPLM2+MLP    & \texttt{vhvl+struct}   & 0.464\scriptsize{$\pm$0.037}  & 0.514\scriptsize{$\pm$0.027}  & 0.125\scriptsize{$\pm$0.111}  & \underline{\textbf{0.132}}\scriptsize{$\pm$0.184} \\
AbMPNN+MLP   & \texttt{struct\_only}  & -0.092\scriptsize{$\pm$0.055} & 0.001\scriptsize{$\pm$0.075}  & 0.120\scriptsize{$\pm$0.103}  & 0.002\scriptsize{$\pm$0.228} \\
\rowcolor{LightCyan}
LinearReg    & \texttt{vhvl\_only}    & 0.531\scriptsize{$\pm$0.044}  & \underline{\textbf{0.714}}\scriptsize{$\pm$0.032} & \underline{\textbf{0.191}}\scriptsize{$\pm$0.255} & 0.035\scriptsize{$\pm$0.283}  \\
\bottomrule
\end{tabular}}
\end{table*}

\begin{table*}[!t]
\centering
\small
\setlength{\tabcolsep}{4pt} 
\renewcommand{\arraystretch}{1.1} 
\caption{Protein yield regression using per-residue 3D coordinate features (\texttt{3D\_coord}). Rows correspond to data splits; columns compare \texttt{LinearReg} vs.\ \texttt{1DCNN} head-to-head for Pearson and Spearman correlations. Values are mean$\pm$std across runs. Best per split/metric is \underline{\textbf{underlined in bold}}.}
\label{tab:appendix-struct-regression}
\begin{tabular}{l
                c c
                c c}
\toprule
\multirow{2}{*}{\textbf{Split}} 
 & \multicolumn{2}{c}{\textbf{Pearson}} 
 & \multicolumn{2}{c}{\textbf{Spearman}} \\
 & \textbf{LinearReg} & \textbf{1DCNN} 
 & \textbf{LinearReg} & \textbf{1DCNN} \\
\midrule
 \texttt{scfv\_signature} 
 & 0.495{\scriptsize$\pm$0.034} & \underline{\textbf{0.529}}{\scriptsize$\pm$0.034}
 & 0.512{\scriptsize$\pm$0.037} & \underline{\textbf{0.546}}{\scriptsize$\pm$0.039} \\
\texttt{Parental\_Family} 
 & 0.007{\scriptsize$\pm$0.167} & \underline{\textbf{0.027}}{\scriptsize$\pm$0.133}
 & 0.002{\scriptsize$\pm$0.114} & \underline{\textbf{0.088}}{\scriptsize$\pm$0.157} \\
\texttt{Fam1} 
 & \underline{\textbf{0.138}}{\scriptsize$\pm$0.106} & 0.070{\scriptsize$\pm$0.252}
 & \underline{\textbf{0.221}}{\scriptsize$\pm$0.134} & 0.176{\scriptsize$\pm$0.417} \\
 \texttt{Fam2} 
 & 0.006{\scriptsize$\pm$0.038} & \underline{\textbf{0.020}}{\scriptsize$\pm$0.075}
 & -0.001{\scriptsize$\pm$0.035} & \underline{\textbf{0.015}}{\scriptsize$\pm$0.071} \\
\texttt{Fam3} 
 & \underline{\textbf{0.187}}{\scriptsize$\pm$0.290} & 0.114{\scriptsize$\pm$0.243}
 & \underline{\textbf{0.022}}{\scriptsize$\pm$0.246} & -0.002{\scriptsize$\pm$0.203} \\
\bottomrule
\end{tabular}
\end{table*}


\end{document}